\documentclass{ecai}  

\usepackage{graphicx}
\usepackage{latexsym}
\usepackage{amsmath} 
\usepackage{amssymb} 
\usepackage{booktabs} 
\usepackage{xcolor} 
\usepackage{multirow} 
\usepackage{array}        
\usepackage{hyperref}

\begin{document}

\begin{frontmatter}

\title{EMTSF: Extraordinary Mixture of SOTA Models for Time Series Forecasting}

\author[A]{\fnms{Musleh}~\snm{Alharthi}\thanks{Corresponding author Email: muslehal@my.bridgeport.edu}}
\author[B]{\fnms{Kaleel}~\snm{Mahmood}}
\author[A]{\fnms{Sarosh}~\snm{Patel}}
\author[A]{\fnms{Ausif}~\snm{Mahmood}}

\address[A]{Computer Science and Engineering, University of Bridgeport, Bridgeport, CT, USA}
\address[B]{Department of Computer Science and Statistics, University of Rhode Island, Kingston, RI, USA}

\begin{abstract}
The immense success of the Transformer architecture in Natural Language Processing has led to its adoption in Time Series Forecasting (TSF), where superior performance has been shown. However, a recent important paper questioned their effectiveness by demonstrating that a simple single layer linear model outperforms Transformer-based models. This was soon shown to be not as valid, by a better transformer-based model termed PatchTST. More recently, TimeLLM demonstrated even better results by repurposing a Large Language Model (LLM) for the TSF domain. Again, a follow up paper challenged this by demonstrating that removing the LLM component or replacing it with a basic attention layer in fact yields better performance. One of the challenges in forecasting is the fact that TSF data favors the more recent past, and is sometimes subject to unpredictable events. Based upon these recent insights in TSF, we propose a strong Mixture of Experts (MoE) framework. Our method combines the state-of-the-art (SOTA) models including xLSTM, enhanced Linear, PatchTST, and minGRU, among others. This set of complimentary and diverse models for TSF are integrated in a Transformer based MoE gating network.  Our proposed model outperforms all existing TSF models on standard benchmarks, surpassing even the latest approaches based on MoE frameworks.

\noindent\textbf{Our code is available at:}  \url{https://github.com/muslehal/EMTSF}
\end{abstract}

\end{frontmatter}

\section{Introduction}

Time Series Forecasting (TSF) models predict future values based on patterns learned from historical data. TSF is used in many different fields such as weather, health care, traffic, finance, electricity consumption, and market demand, among others. TSF has been a challenging area due to the time dependent nature of data which is often susceptible to factors like seasonality, trend changes, uncommon events, non-stationary nature, and noise issues. While TSF has been an active area of research for decades, it has drawn renewed attention due to the advances in the field of AI. Recent research works have explored AI architectures ranging from simple linear neural networks to enhanced transformer-based architectures, as well as state-space based models. As for the best AI-based architecture for TSF, there have been some interesting antithetical developments in the recently published literature. Before we delve into the details of this perplexity, we briefly describe some of the important research in the field of TSF in a chronological manner.

Some of the early successes in the TSF field have utilized classical statistics and mathematics-based approaches involving moving average filters, exponential smoothing, and AutoRegressive Integrated Moving Average (ARIMA) processing. By taking into seasonality patterns, techniques such as SARIMA~\cite{arima_ref} and TBATs~\cite{de2011forecasting} provide enhanced prediction as compared to ARIMA. In the last decade, classical machine learning models such as Linear Regression~\cite{ristanoski2013time}, XGBoost~\cite{chen2016xgboost}, Random Forests and ensemble methods~\cite{masini2023machine} have been explored for the TSF field. These techniques require the data to be transformed into a supervised learning problem by using a sliding window approach. Such models improve performance as they cancel out uncorrelated errors in the averaging process. For larger training data, machine learning approaches usually outperform the classical mathematical techniques of SARIMA and TBATs~\cite{de2011forecasting}. 

The highly successful Convolutional Neural Networks (CNNs) in computer vision were also attempted for TSF. One such example is the Long and Short-term Time-series network (LSTNet) proposed in~\cite{lai2018modeling} which used both the CNN and RNN to extract both short and long-term dependency patterns for improving the TSF. Most of the recent research in the TSF domain uses the Transformer architecture that was originally proposed for Natural Language Processing (NLP)~\cite{vaswani2017attention}. The Transformer uses the attention mechanism to determine the pair-wise similarity in the input sequence to predict the output. It has revolutionized the AI field, leading to development of LLMs such as ChatGPT~\cite{achiam2023gpt}, Llama~\cite{touvron2023llama, dubey2024llama}, Gemini~\cite{team2023gemini}, DeepSeek~\cite{guo2025deepseek}, among others. Thus, it is natural that these developments be applied to create more effective TSF models.   

\section{Preliminaries}
\label{sec:prelims}
In a TSF problem, we are given historical time series data $\textbf{x}$, where $\textbf{x} \in \mathbb{R}^{L \times m}$ 
 with $m$ being the number of multivariate time series features and $L$ the number of look-back time steps. The goal is then to train a model that learns to predict the output $\textbf{y}$,  where  $\textbf{y} \in \mathbb{R}^{T \times m}$ is the prediction for the future $T$ time steps. 
When using a Transformer-based model for TSF, the input data is often divided into a sequence of $n$ patches, i.e., $\mathbf{x_{1}, x_{2}, …., x_{n}}$, where $\mathbf{x_{i}} \in \mathbb{R}^{p \times m}$ with $p=\lfloor \frac{L}{n} \rfloor$. These patches are converted to embedding vectors of size $d\times1$ by a linear transformation according to a fixed dimension $d$ for the transformer model. Position vectors of size  $d\times1$ are added to each embedding vector to maintain relative temporal information.

In the Transformer architecture, three parallel learnt representations of the above transformed data are computed producing $\mathbf{Q}$, $\mathbf{K}$ and $\mathbf{V}$ matrices, with each $\in \mathbb{R}^{n \times d}$. The Transformer then computes a simple similarity in the form of an inner product on the learnt position encoded embeddings of the sequence of $n$ input patches. The pairwise similarity between patches computed as, $\mathbf{A}=\text{softmax}(\mathbf{QK}^{T})$ is referred to as the “attention”. If there are $n$ patches being input, referred to as the context, then $\mathbf{A} \in \mathbb{R}^{n \times n}$. Each layer in the Transformer divides the attention calculation into parallel heads by dividing the data along the embedding dimension. The output from a Transformer layer has the same dimensionality as the input, and is obtained by a matrix computation of $(\mathbf{A} \times \mathbf{V}) \in \mathbb{R}^{n \times d}$  where $\mathbf{V} \in \mathbb{R}^{n \times d}$  contains rows of learnt position encoded representations of the input patches. For a Transformer-based TSF implementation, the output is often produced in one step rather than an autoregressive generation of one patch size at a time.

\section{Related Work}
Some of the important works that use the Transformer architecture for TSF include: Informer~\cite{zhou2021informer}, Autoformer~\cite{wu2021autoformer}, Fedformer~\cite{zhou2022fedformer}, Pyraformer~\cite{liu2022pyraformer},  PatchTST~\cite{Yuqietal-2023-PatchTST} and iTransformer~\cite{liuitransformer}. To better adapt the Transformer for TSF, some pre or post processing features have been added to it, e.g., an auto-correlation mechanism is used in Autoformer and  series decomposition blocks are included in~\cite{wu2021autoformer}.  

Since there is a strong relationship between the time domain and its frequency transformation, Fedformer \cite{zhou2022fedformer} uses a Fourier enhanced structure. To capture long-range dependencies in data, Pyraformer \cite{liu2022pyraformer} builds hierarchical representation of the time series by summarizing features at different resolutions, and modeling temporal dependencies across various scales. PatchTST \cite{Yuqietal-2023-PatchTST} utilizes patching and channel independence to capture dependencies in multivariate time series data. Note that relatively large Transformer models are prone to overfitting in TSF. For example, it has been observed that in the financial time series data, the actual signal is often subtle \cite{zeng2023transformers}, requiring a much smaller model for better prediction.

Some of the Transformer-based models for TSF have demonstrated better results at the time of their writing, e.g., \cite{zhou2021informer,wu2021autoformer,liu2022pyraformer,zhou2022fedformer,liuitransformer}. However, the LTSF-Linear work in \cite{zeng2023transformers} questioned the use of Transformers, reasoning that the permutation-invariant self-attention mechanism may result in temporal information loss. The work in \cite{zeng2023transformers} used an extremely simple one layer linear network producing better forecasting results than the previous transformer-based approaches.

One of the most successful Transformer-based TSF implementation has been PatchTST \cite{Yuqietal-2023-PatchTST}. It segments the time series data into patches, but unlike other models,  channel independence is maintained between variates. 
PatchTST demonstrated significantly superior results as compared to DLinear \cite{zeng2023transformers}, FEDFormer \cite{zhou2022fedformer}, Autoformer \cite{wu2021autoformer} and Informer \cite{zhou2021informer} models. Another important work in adapting the Transformer to the TSF domain is iTransformer \cite{liuitransformer}, which uses the attention mechanism to capture the correlations between different time series (variates) at all historical time points. This allows the model to directly learn how different series influence each other. Since iTransformer does not directly focus on the sequential dependencies within a single time series, this might limit its ability to capture complex, long-range autocorrelations within a series.

The immense success of LLMs in different domains such as NLP, vision, mathematical understanding and reasoning, has raised the question of their viability for the TSF domain. An important work in this context has been recently carried out in \cite{jin2024time}, referred to as the Time-LLM. Here, the authors reprogram the input time series data with text prototypes before feeding it into the frozen LLM to align the two modalities. Their technique referred to as, Prompt-as-Prefix (PaP), directs the transformation of input TSF patches. The resultant output patches from the LLM are used to obtain the forecasts. The reported results in TIME-LLM outperformed the previous SOTA  forecasting models.

Interestingly, the reasons for the success of TIME-LLM were delineated in a follow-up recent work in \cite{tan2024language}. By conducting ablation studies
 on three LLM-based TSF methods (\cite{zhou2023one, liu2024calf,jin2024time}), the authors demonstrated
 that removing the LLM component or replacing it with a basic attention layer yields the same performance, and in most cases, in fact improves performance. Further, computationally expensive pre-trained LLMs do not perform better than models trained from scratch. LLM models also do not represent the sequential dependencies in time series well, and do not assist in few-shot settings. 

 Based on challenges in achieving a single effective model for TSF, more recent research has focused on a Mixture of Experts (MoE) approach, e.g., \cite{ni2024mixture,shi2025time,niu2024mixture,sahoo2024moiraimoe,liang2024moirai,wang2024timemixer}. The work in~\cite{ni2024mixture} argues that even though simple linear transformation based models achieve good results in TSF, due to their inherent simplicity, they are unable to effectively adapt to periodic changes in time series patterns. Thus, they propose an MoE style augmentation for linear-centric models termed Mixture-of-Linear-Experts (MoLE). They train multiple linear-centric models (experts) and a router model that weighs and mixes the expert's outputs. The results reported in MoLE demonstrated some success over individual linear-centric models of DLinear and RLinear \cite{zeng2023transformers}, as well as PatchTST \cite{Yuqietal-2023-PatchTST}.  
 
 A follow up work in~\cite{niu2024mixture} referred to as Mixture of Projection Experts (MoPE) used multiple projection branches instead of a single final projection in a Transformer design to improve the capacity of the network. Similarly, work in~\cite{wang2024timemixer} uses an MLP-based architecture in an MoE design with Past-Decomposable-Mixing (PDM) and Future-Multipredictor-Mixing (FMM) blocks.  This aides in learning disentangled multiscale series in both past and future prediction phases. Another work, termed MOIRAI \cite{liang2024moirai,sahoo2024moiraimoe} is an MoE transformer-based model that uses a different patch size for each granularity and learns a mixture of distributions. It also introduces an elegant attention mechanism that respects permutation variance between each variate to capture the temporal dynamics between data points. While MOIRAI is good for multi-source and multi-resolution data, it may not effectively capture long term time series relationships.

 Following the LLM philosophy, Time-MoE \cite{shi2025time} used a massive data (300 billion time points) approach to train a 2.4 billion parameter LLM designed for TSF. The results surpassed other reported TSF works, however, the complexity of the model and its high training cost are a concern. Further, the Time-300B training dataset which encompasses data from various domains may lead to potential imbalances affecting the model's generalization capabilities.  

 Based on the previously mentioned recent research findings and the challenges in effective TSF modeling, we propose a confluential MoE approach combining  knowledge from a diverse set of complementary strong experts. We  include both linear-based, transformer-based, recently proposed xLSTM based, and minGRU based architectures that we adapt for TSF. We elaborate on these models, and the reasons behind their selection in our MoE framework in the next section. 
 Our contributions can be summarized as:
 \begin{enumerate}
  \item Incorporating strong complimentary experts using linear, transformer, xLSTM and minGRU based approaches that we adapt for TSF in the MoE framework. 
  \item Development of a confluential MoE architecure  where the gating network itself is transformer-based.
 \item Detailed insights via empirical results on various datasets to demonstrate the effectiveness of our MoE approach for TSF.
 \item
 Strong results on standard datasets surpassing existing approaches including recent MoE designs.
 \end{enumerate}

\section{Proposed Method}
\subsection{EMTSF: \textbf{E}xtraordinary \textbf{M}ixture of SOTA Models for \textbf{T}ime \textbf{S}eries \textbf{F}orecasting}

\begin{figure}
    \centering
\includegraphics[width=0.5\textwidth, keepaspectratio]{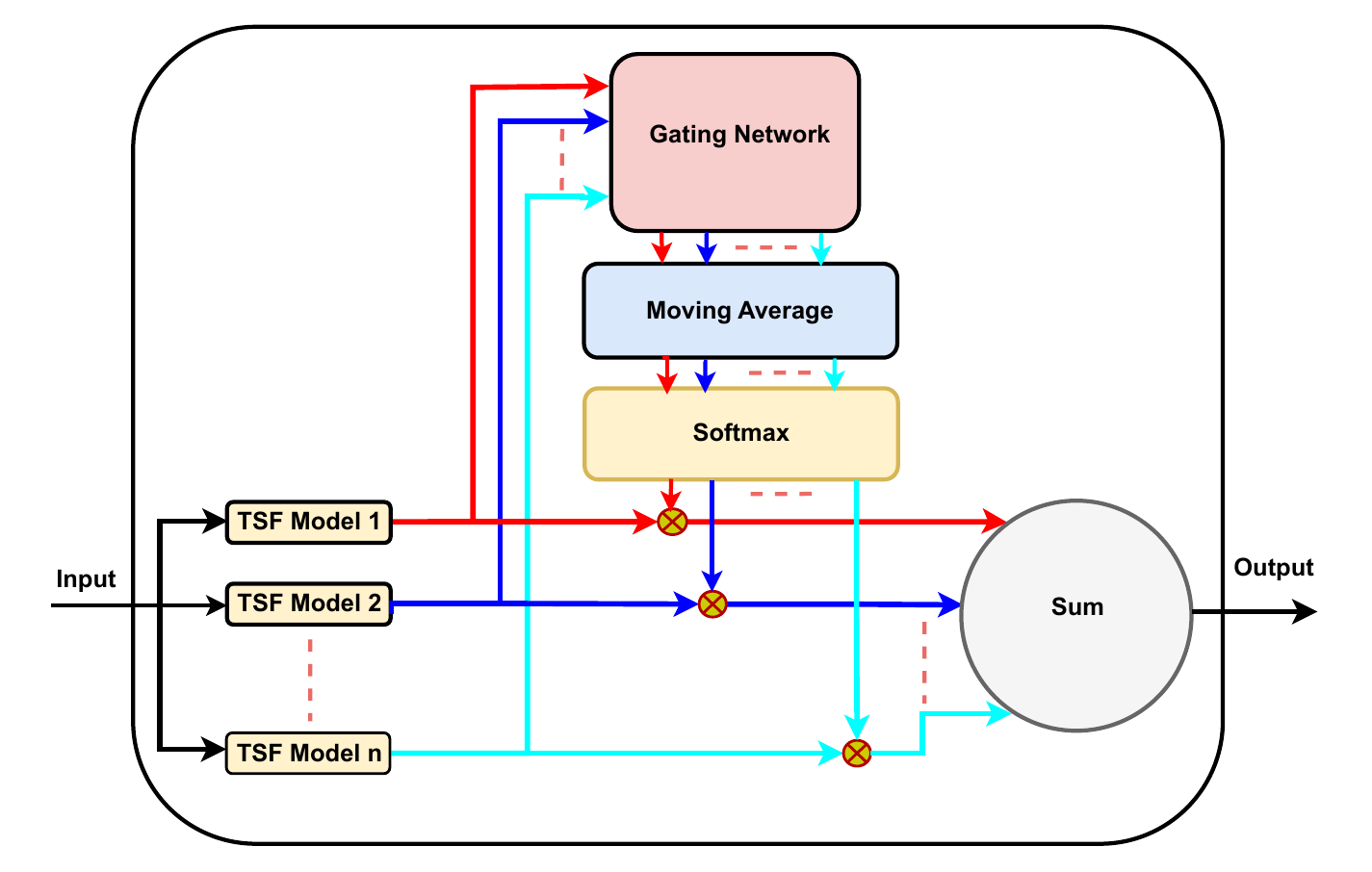}
    \caption{EMTSF: Mixture of Experts Architecture}
    \label{fig:FigureMoE}
\end{figure}

\begin{figure}
    \centering
\includegraphics[width=0.5\textwidth, keepaspectratio]{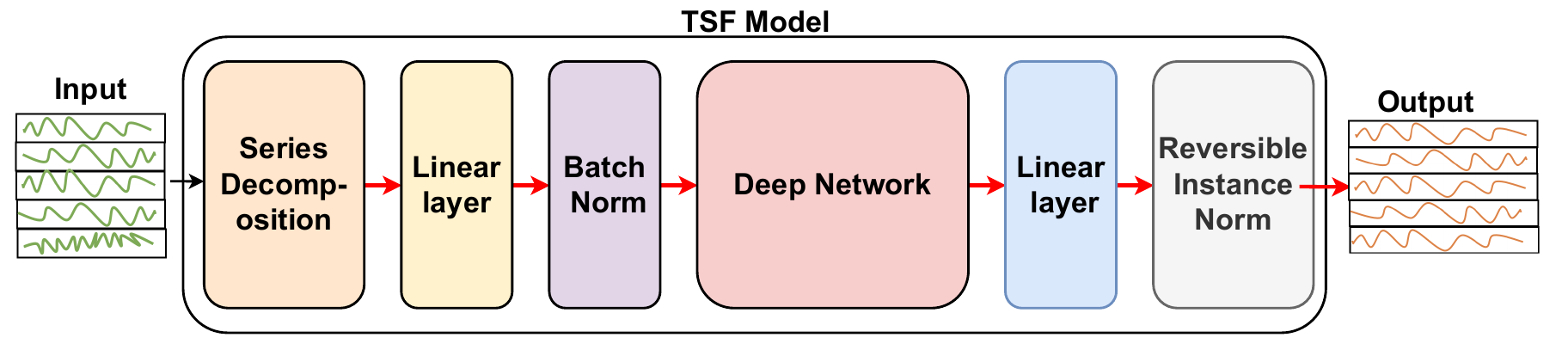}
    \caption{TSF Model with Pre and Post-Processing Blocks}
    \label{fig:TSFModel}
\end{figure}

In our proposed EMTSF architecture, different models for TSF are integrated in a Mixture of Experts (MoE) framework as shown in Figure~\ref{fig:FigureMoE}. The gating network is designed such that it can control the percentage weight of each expert at each time point in the predicted output. In contrast, in a traditional MoE approach, the gating network is usually a simple linear network that outputs a single coefficient for each expert for all time points. We implement both a Transformer-based gating network as well as a simple linear model for it. We further provide smoothing of the coefficient weights from the gating network by a moving average process, as shown in Figure \ref{fig:FigureMoE} and Equations 1-2.
\begin{equation}
g_i(x) = Softmax(\text{${MA_k}$}({G}(cat(TSF Model_i(x)))))_i
\end{equation}
where the function \textit{G} indicates the gating network in MoE (which is itself a transformer network) operating on the sequences of outputs from individual expert TSF models. If the TSF models produce output for \textit{T} timesteps, then the moving average \textit{MA} is carried out for \textit{k} timesteps for each model's gating coefficients where $k \ll T$. This provides smoother combination of different models in the mixture over the range of the predicted timesteps. The final output from the MoE model with $n$ experts is given as:
\begin{equation}
output = \sum_{i=1}^{n} \text{g}_i(x)\, {\text{TSF Model}}_i(x)
\end{equation}
In the TSF domain, it has been well established that pre-processing the data via series decomposition. batch normalization, and post processing via instance normalization yields better prediction results \cite{ni2024mixture,zeng2023transformers}. Thus, each of the experts (TSF models) in the EMTSF utilizes these processing steps as depicted in Figure~\ref{fig:TSFModel}. 
\vspace{0.5cm}
The Series Decomposition block splits the input time series data into two components to capture trend and seasonal information for each series. This approach was proposed in \cite{wu2021autoformer} and is formally described as: For the input sequence with context length of $L$  and $m$ variates, i.e., $\mathbf{x}\in \mathbb{R}^{L \times m}$,  learnable moving averages are applied on each feature via 1-D convolutions. The trend and seasonal components are then extracted as:
\begin{equation}
\mathbf{x}_{trend}=AveragePool(Padding(\mathbf{x}))
\end{equation}
\begin{equation}
\mathbf{x}_{seasonal}=\mathbf{x}- \mathbf{x}_{trend}
\end{equation}
After series decomposition, the data passes through a linear transformation layer to transform it to the dimensionality needed for the deep network for TSF. Batch normalization is further applied to improve stability in learning. The predicted output is transformed back to individual series components via another linear transformation. 

A post processing step of Reversible Instance Normalization (RevIN) \cite{kim2021reversible} has proven to improve the TSF performance \cite{ni2024mixture}.
The RevIn operates on each channel independently. It applies a learnable transformation to normalize the data during training, such that it can be reversed to its original scale during prediction. The loss functions used in the training of a TSF model are the standard $L2$ (MSE) and $L1$ (MAE) losses. 

\subsection{Expert TSF Models in EMTSF}
Our EMTSF framework is generalizable and any individual TSF model can be integrated in it. For our current implementation, we chose four complementary models that have individually proven to be very effective in the TSF domain. The TSF models selected are: PatchTST~\cite{Yuqietal-2023-PatchTST}, Enhanced Linear Model (ELM)~\cite{alharthi2024enhanced}, xLSTMTime~\cite{alharthi2024xlstmtime}, and a new model for TSF that we develop in this work based on the recent minGRU design~\cite{feng2024were}. We term this model for TSF as minGRUTime. From a deep learning architecture stand point, one model in EMTSF is based on the Transformer, another is based on a linear network, and the two others are recent enhanced forms of recurrent models. We briefly describe the design of each one of these in the following subsections.

\subsubsection{PatchTST Model}
PatchTST \cite{Yuqietal-2023-PatchTST} is one of the most successful Transformer-based models for TSF. Its key design includes segmentation of time series into subseries-level patches, and
 channel-independence where each component in the multivariate series is treated as a single univariate time series. The embedding and Transformer weights are shared across all the series. 

 Formally, the $i_{th}$  series for $L$ time steps is treated as a univariate $x_{1:L}^{(i)} = (x_{1}^{(i)}, \ldots, x_{L}^{(i)})$. Each of these is fed independently to the Transformer after converting to patches. The Transformer then learns to predict the output as $\hat{x} = (\hat{x}_{L+1}^{(i)}, \ldots, \hat{x}_{L+T}^{(i)}) \in \mathbb{R}^{1 \times T}$  for the $T$ future steps. For a patch length $P$ and stride $S$,  the patching process generates a sequence of $n$ patches $x_{p}^{(i)} \in \mathbb{R}^{P \times N}$ where $n = \left\lfloor \frac{(L-P)}{S} \right\rfloor + 2$. We chose the PatchTST as one of the TSF models in our EMTSF architecture as the patching design retains local semantic information in the embedding, and the model can effectively attend to the past history due to the attention mechanism of the Transformer.

 \subsubsection{Enhanced Linear Model}
 The success of simple linear network based models (e.g., DLinear and NLinear) in \cite{zeng2023transformers} resulted in their use in subsequent works for TSF e.g., \cite{ni2024mixture}. The work in \cite{alharthi2024enhanced} further used dual pipelines to improve the results of DLinear and NLinear.  Their model is referred to as Enhanced Linear Model (ELM). One of the pipelines in ELM uses DLinear approach while the other uses NLinear. The advantage of NLinear is that it is able to better handle the distribution shift in the data. It does so by subtracting the last value of the sequence, which it adds it back after the linear layer, and before doing the final prediction. 
 
 In ELM, similar to PatchTST, channel independence is maintained. The architecture also utilizes batch normalization and reversible instance normalization \cite{kim2021reversible}. They also used a customized loss function combining the $L1$ and $L2$ losses as:
 \begin{equation}
 Loss = \alpha \times ||y - \hat{y}||_2 + (1 - \alpha)||y - \hat{y}||_1
 \end{equation}
 The motivation for incorporating ELM with dual pipelines of DLinear and NLinear in our EMTSF architecture is that DLinear is particularly useful for time series with trend and seasonality, while NLinear is effective in handling distribution shifts in TSF.
 
 \subsubsection{xLSTMTime Model}
 LSTMs (Long Short-Term Memory) networks, which are an improvement over Recurrent Neural Network (RNN) can efficiently   handle sequential data such as time series, or NLP. With the emergence of Transformers, their use had declined due to slow and unstable training. In a recent research, the LSTM architecture was revised and significantly improved, termed as xLSTM (Extended LSTM) \cite{beck2024xlstm}. xLSTM greatly enhances the traditional LSTM architecture by borrowing some ideas from the Transformer architecture. It also introduces  exponential gating for better normalization and stabilization, and the integration of residual block backbones. These improvements give xLSTM scalability and stability to perform competitively with state-of-the-art Transformers \cite{beck2024xlstm}. xLSTM has a scalar and matrix variant which are termed sLSTM and mLSTM respectively.

 The mLSTM introduces a matrix memory cell along with a covariance update mechanism for key-value pair storage which significantly increases the model's memory capacity. The work in \cite{alharthi2024xlstmtime} termed as xLSTMTime utilized both the sLSTM and the mLSTM designs in applying it to the TSF domain. Their implementation added the pre and post-processing stages as shown in Figure~\ref{fig:TSFModel}. sLSTM architecture is employed for smaller datasets and mLSTM for larger datasets with relatively higher number of covariates in TSF.

Since the TSF prediction generally favors more recent past for short term prediction, the exponential gating in xLSTM is naturally suited for this. In addition, the LSTM capability in it helps with the long term forecasting, making xLSTM a very effective architecture for TSF. For these reasons, xLSTMTime is one of the expert TSF models in our EMTSF architecture.

\subsubsection{minGRUTime Model}
Similar to the renewed LSTM design in xLSTM, the work by \cite{feng2024were} improves traditional Gated Recurrent Unit (GRU). Their improved version is referred to as minGRU (they also proposed minLSTM).
These have fewer parameters and are fully parallelizable during training. It is shown that their performance can rival Transformers.
The operation of the minGRU is described by the following equations:
\begin{equation}
h_{t}=(1-z_{t})\odot h_{t-1}+z_{t}\odot\tilde{h}_{t}
\end{equation}
\begin{equation}
z_{t} = \sigma(\text{Linear}_{d_h}(x_t))
\end{equation}
\begin{equation}
\tilde{h}_{t} = \text{Linear}_{d_h}(x_t)
\end{equation}
where $\tilde{h}_{t}$ represents the candidate hidden state, a potential new value for the hidden state, $z_t \in (0, 1)$ determines how much of the past information should be carried forward.

Because of the much simpler design in minGRU, and since TSF domain has shown to be less beneficial from more complex models, we incorporate the minGRU design in our EMTSF architecture. The pre and post processing stages are added to the minGRU as indicated in Figure ~\ref{fig:TSFModel}.

Thus overall, we have selected the most promising TSF models based on diverse architectural foundations in our extraordinary mixture of experts (EMTSF). We further use a learnable end-to-end MoE design with a Transformer-based gating network (as indicated in Figure ~\ref{fig:FigureMoE}). This allows the MoE to extract the best performance by combining the models in a complementary manner. The results in the next section highlight the impressive performance of our EMTSF design over the individual experts including existing state of the art TSF models, as well as recent MoE based TSF designs.

\section{Results}
We test our EMTSF  architecture on different datasets. These include the PEMS traffic dataset (from California Transportation Agencies) which can be used for traffic flow prediction, congestion detection, and travel time estimation. We use the PEMS03, PEMS04, PEMS07 and PEMS08 sets from the above dataset. In addition, we analyze our models on 11 widely used datasets from real-world applications that include the Electricity Transformer Temperature (ETT) series. These are further divided into ETTh1 and ETTh2 (hourly intervals), and ETTm1 and ETTm2 (5-minute intervals). We also perform tests on datasets related to Traffic (hourly), Electricity (hourly), Weather (10-minute intervals), and Influenza-Like Illness (ILI) (weekly). The characteristics of the datasets used in our experiments are detailed in Table \ref{tab:datasets}
\begin{table}[ht]
\centering
\renewcommand{\arraystretch}{1.2} 

\begin{tabular}{lccc}
\toprule
\textbf{Datasets} & \textbf{Timesteps} & \textbf{Features} & \textbf{Granularity} \\
\midrule
PEMS03      & 26,209 & 358 & 5 min  \\
PEMS04      & 16,992 & 307 & 5 min  \\
PEMS07      & 28,224 & 883 & 5 min  \\
PEMS08      & 17,856 & 170 & 5 min  \\
Weather     & 52,696 & 21  & 10 min \\
Traffic     & 17,544 & 862 & 1 h    \\
Electricity & 26,304 & 321 & 1 h    \\
Illness     & 966    & 7   & 1 week \\
ETTh1/ETTh2 & 17,420 & 7   & 1 h    \\
ETTm1/ETTm2 & 69,680 & 7   & 5 min  \\
\bottomrule
\end{tabular}
\caption{Characteristics of Time Series Datasets}
\label{tab:datasets}
\end{table}

Table \ref{tab:DatasetsComparisonOtherModels} shows comparison of our EMTSF model with other state-of-the-art (SOTA) models for TSF on the popular datasets. The evaluation metrics used are MSE (Mean Squared Error) and MAE (Mean Absolute Error). As it can be seen that in a vast majority of the cases, our EMTSF model outperforms current SOTA designs. The SOTA models compared in Table \ref{tab:DatasetsComparisonOtherModels} include: TimeLLM \cite{jin2024time}, GPT4TS \cite{zhou2023one}, DLinear \cite{zeng2023transformers}, PatchTST \cite{Yuqietal-2023-PatchTST}, TimesNet \cite{wu2022timesnet}, FEDFormer \cite{zhou2022fedformer}, Autoformer \cite{wu2021autoformer}, Stationary \cite{liu2022non} and ETSformer \cite{woo2022etsformer}. The look back window size in Table \ref{tab:DatasetsComparisonOtherModels} is $L$ =512 for all datasets except for ILI which uses $L$=96. The predicted length $T$ varies with values of \{96,192,336,720\} and is listed under the Horizon column. For the ILI dataset, the predicted length is different as shown in Table \ref{tab:DatasetsComparisonOtherModels}. The red color indicates the best result while the blue indicates the second best for a given category. As can be seen, our EMTS deign produces the best MSE and MAE results in majority of the cases. In one category, PatchTST performs the best (Etth2 with a prediction length of 336). In a few cases, TimeLLM has better results. Overall, TimeLLM is second best amongst the models compared in Table  \ref{tab:DatasetsComparisonOtherModels}. 

Table \ref{tab:ComparisonTimeMixeriTrans} shows the comparison of our EMTSF model with TimeMixer++, TimeMixer \cite{wang2024timemixer} and iTransformer \cite{liuitransformer} models. Here also, our EMTSF model outperforms the TimeMixer and iTransformer models. In almost all categories, our model produces best results (lowest MSE and MAE values), followed by TimeMixer++ as the second best model.

Our EMTSF design with four expert models in the mixture uses only 9.66 million parameters. In comparison, Time-LLM \cite{jin2024time} is based on a 6.6 billion parameter LLM, while the Time-MOE \cite{shi2025time} uses 2.4 billion parameters.

\begin{table*}[t] 
\centering
\scalebox{0.70}{

\begin{tabular}{|c|c|c|c|c|c|c|c|c|c|c|c|}
\hline
\textbf{Dataset} & \textbf{Horizon} & \textbf{EMTSF(ours)} & \textbf{TIME-LLM} & \textbf{GPT4TS} & \textbf{DLinear} & \textbf{PatchTST} & \textbf{TimesNet} & \textbf{FEDformer} & \textbf{Autoformer} & \textbf{Stationary} & \textbf{ETSformer}  \\ \hline
\textbf{Metric} &  & \textbf{MSE MAE} & \textbf{MSE MAE} & \textbf{MSE MAE} & \textbf{MSE MAE} & \textbf{MSE MAE} & \textbf{MSE MAE} & \textbf{MSE MAE} & \textbf{MSE MAE} & \textbf{MSE MAE} & \textbf{MSE MAE} \\ \hline 

\multirow{4}{*}{\textbf{ETTh1}} & 96 &\textcolor{red}{ 0.359\hspace{0.5cm}0.384} & \textcolor{blue}{0.362\hspace{0.5cm}0.392} & 0.376\hspace{0.5cm}0.397 & 0.375\hspace{0.5cm}0.399 & 0.370\hspace{0.5cm}0.399 & 0.384\hspace{0.5cm}0.402 & 0.376\hspace{0.5cm}0.419 & 0.449\hspace{0.5cm}0.459 & 0.513\hspace{0.5cm}0.491 & 0.494\hspace{0.5cm}0.479 \\ 
 & 192 &\textcolor{blue}{ 0.399}\hspace{0.5cm}\textcolor{red}{0.411} &\textcolor{red}{ 0.398}\hspace{0.5cm}\textcolor{blue}{0.418} & 0.416\hspace{0.5cm}0.418 & 0.405\hspace{0.5cm}0.416 & 0.413\hspace{0.5cm}0.421 & 0.436\hspace{0.5cm}0.429 & 0.420\hspace{0.5cm}0.448 & 0.500\hspace{0.5cm}0.482 & 0.534\hspace{0.5cm}0.504 & 0.538\hspace{0.5cm}0.504 \\  
 & 336 & \textcolor{red}{0.418\hspace{0.5cm}0.422} &\textcolor{blue}{ 0.430\hspace{0.5cm}0.427} & 0.442\hspace{0.5cm}0.433 & 0.439\hspace{0.5cm}0.443 & 0.422\hspace{0.5cm}0.436 & 0.491\hspace{0.5cm}0.469 & 0.459\hspace{0.5cm}0.465 & 0.521\hspace{0.5cm}0.496 & 0.588\hspace{0.5cm}0.535 & 0.574\hspace{0.5cm}0.521 \\ 
 & 720 &\textcolor{red}{ 0.436\hspace{0.5cm}0.454} & \textcolor{blue}{0.442\hspace{0.5cm}0.457 }& 0.477\hspace{0.5cm}0.456 & 0.472\hspace{0.5cm}0.490 & 0.447\hspace{0.5cm}0.466 & 0.521\hspace{0.5cm}0.500 & 0.506\hspace{0.5cm}0.507 & 0.514\hspace{0.5cm}0.512 & 0.643\hspace{0.5cm}0.616 & 0.562\hspace{0.5cm}0.535 \\
 & Avg & \textcolor{red}{0.403\hspace{0.5cm}0.417} & \textcolor{blue}{0.408\hspace{0.5cm}0.423} & 0.422\hspace{0.5cm}0.437 & 0.413\hspace{0.5cm}0.430 & 0.458\hspace{0.5cm}0.450 & 0.440\hspace{0.5cm}0.460 & 0.496\hspace{0.5cm}0.487 & 0.570\hspace{0.5cm}0.537 & 0.542\hspace{0.5cm}0.510 & 0.491\hspace{0.5cm}0.479 \\ \hline

\multirow{4}{*}{\textbf{ETTh2}} & 96 &\textcolor{red}{ 0.262\hspace{0.5cm}0.324} & \textcolor{blue}{0.268\hspace{0.5cm}0.328} & 0.285\hspace{0.5cm}0.342 & 0.289\hspace{0.5cm}0.353 & 0.274\hspace{0.5cm}0.336 & 0.340\hspace{0.5cm}0.374 & 0.358\hspace{0.5cm}0.397 & 0.346\hspace{0.5cm}0.388 & 0.476\hspace{0.5cm}0.458 & 0.340\hspace{0.5cm}0.391 \\ 
 & 192 &\textcolor{red}{ 0.328\hspace{0.5cm}0.371} & \textcolor{blue}{0.329\hspace{0.5cm}0.375} & 0.354\hspace{0.5cm}0.389 & 0.383\hspace{0.5cm}0.418 & 0.339\hspace{0.5cm}0.379 & 0.402\hspace{0.5cm}0.414 & 0.429\hspace{0.5cm}0.439 & 0.456\hspace{0.5cm}0.452 & 0.512\hspace{0.5cm}0.493 & 0.430\hspace{0.5cm}0.439 \\  
 & 336 &\textcolor{blue}{ 0.347\hspace{0.5cm}0.387} & 0.368\hspace{0.5cm}0.409 & 0.373\hspace{0.5cm}0.407 & 0.448\hspace{0.5cm}0.465 & \textcolor{red}{0.329\hspace{0.5cm}0.380} & 0.452\hspace{0.5cm}0.452 & 0.496\hspace{0.5cm}0.487 & 0.482\hspace{0.5cm}0.486 & 0.552\hspace{0.5cm}0.551 & 0.485\hspace{0.5cm}0.479 \\ 
 & 720 & 0.381\hspace{0.5cm}\textcolor{red}{0.417 }& \textcolor{red}{0.372}\hspace{0.5cm}\textcolor{blue}{0.420} & 0.406\hspace{0.5cm}0.441 & 0.605\hspace{0.5cm}0.551 & \textcolor{blue}{ 0.379}\hspace{0.5cm}0.422 & 0.462\hspace{0.5cm}0.468 & 0.463\hspace{0.5cm}0.474 & 0.515\hspace{0.5cm}0.511 & 0.562\hspace{0.5cm}0.560 & 0.500\hspace{0.5cm}0.497 \\ 
 & Avg & \textcolor{red}{0.329\hspace{0.5cm}0.374} & 0.334\hspace{0.5cm}0.383 & 0.381\hspace{0.5cm}0.412 & 0.431\hspace{0.5cm}0.446 & \textcolor{blue}{0.330\hspace{0.5cm}0.379} & 0.414\hspace{0.5cm}0.427 & 0.437\hspace{0.5cm}0.449 & 0.450\hspace{0.5cm}0.459 & 0.526\hspace{0.5cm}0.516 & 0.439\hspace{0.5cm}0.452 \\ \hline
 
\multirow{4}{*}{\textbf{ETTm1}} & 96 & \textcolor{red}{0.271\hspace{0.5cm}0.325} & \textcolor{blue}{0.272\hspace{0.5cm}0.334} & 0.292\hspace{0.5cm}0.346 & 0.299\hspace{0.5cm}0.343 & 0.290\hspace{0.5cm}0.342 & 0.338\hspace{0.5cm}0.375 & 0.379\hspace{0.5cm}0.419 & 0.505\hspace{0.5cm}0.475 & 0.386\hspace{0.5cm}0.398 & 0.375\hspace{0.5cm}0.398 \\ 
 & 192 &\textcolor{blue}{ 0.322}\hspace{0.5cm}\textcolor{red}{0.351} &\textcolor{red}{ 0.310}\hspace{0.5cm}\textcolor{blue}{0.358 }& 0.332\hspace{0.5cm}0.372 & 0.335\hspace{0.5cm}0.365 & 0.332\hspace{0.5cm}0.369 & 0.374\hspace{0.5cm}0.387 & 0.426\hspace{0.5cm}0.441 & 0.553\hspace{0.5cm}0.496 & 0.459\hspace{0.5cm}0.444 & 0.408\hspace{0.5cm}0.410 \\  
 & 336 & \textcolor{red}{0.350\hspace{0.5cm}0.370} &\textcolor{blue}{ 0.352\hspace{0.5cm}0.384} & 0.366\hspace{0.5cm}0.394 & 0.369\hspace{0.5cm}0.386 & 0.366\hspace{0.5cm}0.392 & 0.410\hspace{0.5cm}0.411 & 0.445\hspace{0.5cm}0.459 & 0.621\hspace{0.5cm}0.537 & 0.495\hspace{0.5cm}0.464 & 0.435\hspace{0.5cm}0.428 \\ 
 & 720 & \textcolor{blue}{0.414}\hspace{0.5cm}\textcolor{red}{0.404} & \textcolor{red}{0.383}\hspace{0.5cm}\textcolor{blue}{0.411} & 0.417\hspace{0.5cm}0.421 & 0.425\hspace{0.5cm}0.421 & 0.416\hspace{0.5cm}0.420 & 0.478\hspace{0.5cm}0.450 & 0.543\hspace{0.5cm}0.490 & 0.671\hspace{0.5cm}0.561 & 0.585\hspace{0.5cm}0.516 & 0.499\hspace{0.5cm}0.462 \\ 
& Avg & \textcolor{blue}{0.339}\hspace{0.5cm}\textcolor{red}{0.362} & \textcolor{red}{0.329}\hspace{0.5cm}\textcolor{blue}{0.372}  & 0.388\hspace{0.5cm}0.403 & 0.357\hspace{0.5cm}0.378 & 0.351\hspace{0.5cm}0.380 & 0.400\hspace{0.5cm}0.406 & 0.448\hspace{0.5cm}0.452 & 0.588\hspace{0.5cm}0.517 & 0.481\hspace{0.5cm}0.456 & 0.429\hspace{0.5cm}0.425 \\ \hline

\multirow{4}{*}{\textbf{ETTm2}} & 96 & \textcolor{red}{0.156\hspace{0.5cm}0.240} & \textcolor{blue}{0.161\hspace{0.5cm}0.253} & 0.173\hspace{0.5cm}0.262 & 0.167\hspace{0.5cm}0.269 & 0.165\hspace{0.5cm}0.255 & 0.187\hspace{0.5cm}0.267 & 0.203\hspace{0.5cm}0.287 & 0.255\hspace{0.5cm}0.339 & 0.192\hspace{0.5cm}0.274 & 0.189\hspace{0.5cm}0.280 \\ 
 & 192 & \textcolor{red}{0.212\hspace{0.5cm}0.280} & \textcolor{blue}{0.219\hspace{0.5cm}0.293} & 0.229\hspace{0.5cm}0.301 & 0.224\hspace{0.5cm}0.303 & 0.220\hspace{0.5cm}0.292 & 0.249\hspace{0.5cm}0.309 & 0.269\hspace{0.5cm}0.328 & 0.281\hspace{0.5cm}0.340 & 0.280\hspace{0.5cm}0.339 & 0.253\hspace{0.5cm}0.319 \\  
 & 336 & \textcolor{red}{0.263\hspace{0.5cm}0.315} & \textcolor{blue}{0.271\hspace{0.5cm}0.329} & 0.286\hspace{0.5cm}0.341 & 0.281\hspace{0.5cm}0.342 & 0.274\hspace{0.5cm}0.329 & 0.321\hspace{0.5cm}0.351 & 0.325\hspace{0.5cm}0.366 & 0.339\hspace{0.5cm}0.372 & 0.334\hspace{0.5cm}0.361 & 0.314\hspace{0.5cm}0.357 \\ 
 & 720 & \textcolor{red}{0.351\hspace{0.5cm}0.371} &\textcolor{blue}{ 0.352\hspace{0.5cm}0.379} & 0.378\hspace{0.5cm}0.401 & 0.397\hspace{0.5cm}0.421 & 0.362\hspace{0.5cm}0.385 & 0.408\hspace{0.5cm}0.403 & 0.421\hspace{0.5cm}0.415 & 0.433\hspace{0.5cm}0.432 & 0.417\hspace{0.5cm}0.413 & 0.414\hspace{0.5cm}0.413 \\ 
 & Avg & \textcolor{red}{0.245\hspace{0.5cm}0.301} & \textcolor{blue}{0.251\hspace{0.5cm}0.313} & 0.284\hspace{0.5cm}0.339 & 0.267\hspace{0.5cm}0.333 & 0.255\hspace{0.5cm}0.315 & 0.291\hspace{0.5cm}0.333 & 0.305\hspace{0.5cm}0.349 & 0.327\hspace{0.5cm}0.371 & 0.306\hspace{0.5cm}0.347 & 0.293\hspace{0.5cm}0.342 \\ \hline
 
 \multirow{4}{*}{\textbf{Weather}} 
& 96 & \textcolor{red}{0.138\hspace{0.5cm}0.177} & \textcolor{blue}{0.147\hspace{0.5cm}0.201} & 0.162\hspace{0.5cm}0.212 & 0.176\hspace{0.5cm}0.237 & 0.149\hspace{0.5cm}0.198 & 0.172\hspace{0.5cm}0.220 & 0.217\hspace{0.5cm}0.296 & 0.266\hspace{0.5cm}0.336 & 0.173\hspace{0.5cm}0.223 & 0.197\hspace{0.5cm}0.281 \\ 
& 192 & \textcolor{red}{0.181}\hspace{0.5cm}\textcolor{red}{0.220}& \textcolor{blue}{0.189\hspace{0.5cm}0.234} & 0.204\hspace{0.5cm}0.248 & 0.220\hspace{0.5cm}0.282 & 0.194\hspace{0.5cm}0.241 & 0.219\hspace{0.5cm}0.261 & 0.276\hspace{0.5cm}0.336 & 0.307\hspace{0.5cm}0.367 & 0.245\hspace{0.5cm}0.285 & 0.237\hspace{0.5cm}0.312 \\  
& 336 & \textcolor{red}{0.230\hspace{0.5cm}0.260} &\textcolor{blue}{ 0.262\hspace{0.5cm}0.279} & 0.254\hspace{0.5cm}0.286 & 0.265\hspace{0.5cm}0.319 & 0.245\hspace{0.5cm}0.282 & 0.280\hspace{0.5cm}0.306 & 0.339\hspace{0.5cm}0.380 & 0.359\hspace{0.5cm}0.395 & 0.321\hspace{0.5cm}0.338 & 0.298\hspace{0.5cm}0.353 \\  
& 720 & \textcolor{red}{0.304\hspace{0.5cm}0.315} &\textcolor{red}{ 0.304}\hspace{0.5cm} \textcolor{blue}{0.316} & 0.326\hspace{0.5cm}0.337 & 0.333\hspace{0.5cm}0.362 & 0.314\hspace{0.5cm}0.334 & 0.365\hspace{0.5cm}0.359 & 0.403\hspace{0.5cm}0.428 & 0.419\hspace{0.5cm}0.428 & 0.414\hspace{0.5cm}0.410 & 0.352\hspace{0.5cm}0.386 \\ 
& Avg & \textcolor{red}{0.213\hspace{0.5cm}0.243} & \textcolor{blue}{0.225\hspace{0.5cm}0.257} & 0.237\hspace{0.5cm}0.270 & 0.248\hspace{0.5cm}0.300 & 0.225\hspace{0.5cm}0.264 & 0.259\hspace{0.5cm}0.287 & 0.309\hspace{0.5cm}0.360 & 0.338\hspace{0.5cm}0.382 & 0.288\hspace{0.5cm}0.314 & 0.271\hspace{0.5cm}0.334 \\ \hline

\multirow{4}{*}{\textbf{Electricity}} & 96 & \textcolor{red}{0.126\hspace{0.5cm}0.217} & 0.131\hspace{0.5cm}0.224 & 0.139\hspace{0.5cm}0.238 & 0.140\hspace{0.5cm}0.237 & \textcolor{blue}{0.129\hspace{0.5cm}0.222} & 0.168\hspace{0.5cm}0.272 & 0.193\hspace{0.5cm}0.308 & 0.201\hspace{0.5cm}0.317 & 0.169\hspace{0.5cm}0.273 & 0.187\hspace{0.5cm}0.304 \\ 
 & 192 & \textcolor{red}{0.144\hspace{0.5cm}0.234} & \textcolor{blue}{0.152}\hspace{0.5cm}0.241 & 0.153\hspace{0.5cm}0.251 & 0.153\hspace{0.5cm}0.249 & 0.157\hspace{0.5cm}\textcolor{blue}{0.240} & 0.184\hspace{0.5cm}0.289 & 0.201\hspace{0.5cm}0.315 & 0.222\hspace{0.5cm}0.334 & 0.182\hspace{0.5cm}0.286 & 0.199\hspace{0.5cm}0.315 \\  
 & 336 & \textcolor{red}{0.158\hspace{0.5cm}0.248} & \textcolor{blue}{0.160}\hspace{0.5cm}\textcolor{red}{0.248} & 0.169\hspace{0.5cm}0.266 & 0.169\hspace{0.5cm}0.267 & 0.163\hspace{0.5cm}\textcolor{blue}{0.259} & 0.198\hspace{0.5cm}0.300 & 0.214\hspace{0.5cm}0.329 & 0.231\hspace{0.5cm}0.338 & 0.200\hspace{0.5cm}0.304 & 0.212\hspace{0.5cm}0.329 \\ 
 & 720 &\textcolor{red}{ 0.190\hspace{0.5cm}0.277} &\textcolor{blue}{0.192}\hspace{0.5cm}0.298 & 0.206\hspace{0.5cm}0.297 & 0.203\hspace{0.5cm}0.301 & 0.197\hspace{0.5cm}\textcolor{blue}{0.290} & 0.220\hspace{0.5cm}0.320 & 0.246\hspace{0.5cm}0.355 & 0.254\hspace{0.5cm}0.361 & 0.222\hspace{0.5cm}0.321 & 0.233\hspace{0.5cm}0.345 \\ 
 & Avg & \textcolor{red}{0.154\hspace{0.5cm}0.244} & \textcolor{blue}{0.158\hspace{0.5cm}0.252} & 0.167\hspace{0.5cm}0.263 & 0.166\hspace{0.5cm}0.263 & 0.161\hspace{0.5cm}\textcolor{blue}{0.252} & 0.192\hspace{0.5cm}0.295 & 0.214\hspace{0.5cm}0.327 & 0.227\hspace{0.5cm}0.338 & 0.193\hspace{0.5cm}0.296 & 0.208\hspace{0.5cm}0.323 \\ \hline

\multirow{4}{*}{\textbf{Traffic}} 
& 96 & \textcolor{red}{0.343\hspace{0.5cm}0.225} & 0.362\hspace{0.5cm}\textcolor{blue}{0.248} & 0.388\hspace{0.5cm}0.282 & 0.410\hspace{0.5cm}0.282 &\textcolor{blue}{ 0.360}\hspace{0.5cm}0.249 & 0.593\hspace{0.5cm}0.321 & 0.587\hspace{0.5cm}0.366 & 0.613\hspace{0.5cm}0.388 & 0.612\hspace{0.5cm}0.338 & 0.607\hspace{0.5cm}0.392 \\ 
& 192 & \textcolor{red}{0.369\hspace{0.5cm}0.238} & \textcolor{blue}{0.374\hspace{0.5cm}0.247} & 0.407\hspace{0.5cm}0.290 & 0.423\hspace{0.5cm}0.287 & 0.379\hspace{0.5cm}0.256 & 0.617\hspace{0.5cm}0.336 & 0.604\hspace{0.5cm}0.373 & 0.616\hspace{0.5cm}0.382 & 0.613\hspace{0.5cm}0.340 & 0.621\hspace{0.5cm}0.399 \\ 
& 336 & \textcolor{red}{0.382\hspace{0.5cm}0.242} & \textcolor{blue}{0.385}\hspace{0.5cm}0.271 & 0.412\hspace{0.5cm}0.294 & 0.436\hspace{0.5cm}0.296 & 0.392\hspace{0.5cm}\textcolor{blue}{0.264} & 0.629\hspace{0.5cm}0.336 & 0.621\hspace{0.5cm}0.383 & 0.622\hspace{0.5cm}0.337 & 0.618\hspace{0.5cm}0.328 & 0.622\hspace{0.5cm}0.396 \\ 
& 720 & \textcolor{red}{0.424\hspace{0.5cm}0.270} & \textcolor{blue}{0.430}\hspace{0.5cm}0.288 & 0.450\hspace{0.5cm}0.312 & 0.466\hspace{0.5cm}0.315 & 0.432\hspace{0.5cm}\textcolor{blue}{0.286} & 0.640\hspace{0.5cm}0.350 & 0.626\hspace{0.5cm}0.382 & 0.660\hspace{0.5cm}0.408 & 0.653\hspace{0.5cm}0.355 & 0.632\hspace{0.5cm}0.396 \\ 
& Avg & \textcolor{red}{0.379\hspace{0.5cm}0.243} &\textcolor{blue}{ 0.388}\hspace{0.5cm}0.264 & 0.414\hspace{0.5cm}0.294 & 0.433\hspace{0.5cm}0.295 & 0.390\hspace{0.5cm}\textcolor{blue}{0.263} & 0.620\hspace{0.5cm}0.336 & 0.610\hspace{0.5cm}0.376 & 0.628\hspace{0.5cm}0.379 & 0.624\hspace{0.5cm}0.340 & 0.621\hspace{0.5cm}0.396 \\ \hline
\multirow{4}{*}{\textbf{ILI}} & 24 & 1.617\hspace{0.5cm}\textcolor{blue}{0.732} & \textcolor{red}{1.285}\hspace{0.5cm}\textcolor{red}{0.727} & 2.063\hspace{0.5cm}0.881 & 2.215\hspace{0.5cm}1.081 & \textcolor{blue}{1.319}\hspace{0.5cm}0.754 & 2.317\hspace{0.5cm}0.934 & 3.228\hspace{0.5cm}1.260 & 3.483\hspace{0.5cm}1.287 & 2.294\hspace{0.5cm}0.945 & 2.527\hspace{0.5cm}1.020 \\  
 & 36 & 1.586\hspace{0.5cm}\textcolor{red}{0.728} & \textcolor{red}{1.404}\hspace{0.5cm}\textcolor{blue}{0.814} & 1.868\hspace{0.5cm}0.892 & 1.963\hspace{0.5cm}0.963 & \textcolor{blue}{1.430}\hspace{0.5cm}0.834 & 1.972\hspace{0.5cm}0.920 & 2.679\hspace{0.5cm}1.080 & 3.103\hspace{0.5cm}1.148 & 1.825\hspace{0.5cm}0.848 & 2.615\hspace{0.5cm}1.007 \\
 & 48 & 1.587\hspace{0.5cm} \textcolor{red}{0.753} & \textcolor{red}{1.523}\hspace{0.5cm}\textcolor{blue}{0.807} & 1.790\hspace{0.5cm}0.884 & 2.130\hspace{0.5cm}1.024 & \textcolor{blue}{1.553}\hspace{0.5cm}0.815 & 2.238\hspace{0.5cm}0.940 & 2.622\hspace{0.5cm}1.078 & 2.669\hspace{0.5cm}1.085 & 2.010\hspace{0.5cm}0.900 & 2.359\hspace{0.5cm}0.972 \\  
 & 60 & 1.560\hspace{0.5cm}\textcolor{red}{0.768} & \textcolor{blue}{1.531}\hspace{0.5cm}0.854 & 1.979\hspace{0.5cm}0.957 & 2.368\hspace{0.5cm}1.096 &\textcolor{red}{ 1.470}\hspace{0.5cm}\textcolor{blue}{0.788} & 2.027\hspace{0.5cm}0.928 & 2.857\hspace{0.5cm}1.157 & 2.770\hspace{0.5cm}1.125 & 2.178\hspace{0.5cm}0.963 & 2.487\hspace{0.5cm}1.016 \\ 
 & Avg & 1.587\hspace{0.5cm}\textcolor{red}{0.745}& \textcolor{red}{1.435}\hspace{0.5cm}0.801 & 1.925\hspace{0.5cm}0.903 & 2.169\hspace{0.5cm}1.041 & \textcolor{blue}{1.443\hspace{0.5cm}0.797} & 2.139\hspace{0.5cm}0.931 & 2.847\hspace{0.5cm}1.144 & 3.006\hspace{0.5cm}1.161 & 2.077\hspace{0.5cm}0.914 & 2.497\hspace{0.5cm}1.004 \\ \hline
\end{tabular}%
} 
\caption{Performance Comparison of our EMTSF model with other SOTA Models on popular TSF Datasets.}
\label{tab:DatasetsComparisonOtherModels}
\end{table*}

Table \ref{tab:ComparisonMOIRAITimeMOE} shows the comparison of our EMTSF model with Time-MOE \cite{shi2025time} and MOIRAI-MOE \cite{sahoo2024moiraimoe} TSF models. We obtain the best results in most of the cases, with Time-MOE base and large models performing better on the ETTh1 dataset. Overall, as indicated by the comparison results in Tables \ref{tab:DatasetsComparisonOtherModels},\ref{tab:ComparisonTimeMixeriTrans} and \ref{tab:ComparisonMOIRAITimeMOE}, our EMTSF model outperforms the existing SOTA models for TSF including those that are based on the MoE design such as Time-MOE \cite{shi2025time}.

\begin{table}[h!]
\centering

\scalebox{0.75}{ 
\footnotesize 
\setlength{\tabcolsep}{4pt} 
\begin{tabular}{|c|c|cc|cc|cc|cc|}
\hline
\textbf{Dataset} & \textbf{Horizon}
& \multicolumn{2}{c|}{\textbf{EMTSF (Ours)}}
& \multicolumn{2}{c|}{\textbf{TimeMixer++}}
& \multicolumn{2}{c|}{\textbf{TimeMixer (2024b)}}
& \multicolumn{2}{c|}{\textbf{iTransformer (2024)}} \\
\cline{3-10}
& & \textbf{MSE} & \textbf{MAE} & \textbf{MSE} & \textbf{MAE} & \textbf{MSE} & \textbf{MAE} & \textbf{MSE} & \textbf{MAE} \\
\hline
\multirow{5}{*}{ETTh1}
& 96  & \textcolor{red}{0.359} & \textcolor{red}{0.384} & \textcolor{blue}{0.361} & 0.403 & 0.375 & \textcolor{blue}{0.400} & 0.386 & 0.405 \\
& 192  & \textcolor{red}{0.399} & \textcolor{red}{0.411} & \textcolor{blue}{0.416} & 0.441 & 0.429 & \textcolor{blue}{0.421} & 0.441 & 0.512 \\
& 336  & \textcolor{red}{0.418} & \textcolor{red}{0.422} & \textcolor{blue}{0.430} & \textcolor{blue}{0.434} & 0.484 & 0.458 & 0.487 & 0.458 \\
& 720  & \textcolor{red}{0.436} & \textcolor{blue}{0.454} & \textcolor{blue}{0.467} & \textcolor{red}{0.451} & 0.498 & 0.482 & 0.503 & 0.491 \\
& Avg  & \textcolor{red}{0.403} & \textcolor{red}{0.417} & \textcolor{blue}{0.419} & \textcolor{blue}{0.432} & 0.447 & 0.440 & 0.454 & 0.447 \\
\hline
\multirow{5}{*}{ETTh2}
& 96  & \textcolor{red}{0.262} & \textcolor{red}{0.324} & \textcolor{blue}{0.276} & \textcolor{blue}{0.328} & 0.289 & 0.341 & 0.297 & 0.349 \\
& 192  & \textcolor{red}{0.328} & \textcolor{red}{0.371} & \textcolor{blue}{0.342} & \textcolor{blue}{0.379} & 0.372 & 0.392 & 0.380 & 0.400 \\
& 336  & \textcolor{blue}{0.347} & \textcolor{red}{0.387} & \textcolor{red}{0.346} & \textcolor{blue}{0.398} & 0.386 & 0.414 & 0.428 & 0.432 \\
& 720  & \textcolor{red}{0.381} & \textcolor{blue}{0.417} & \textcolor{blue}{0.392} & \textcolor{red}{0.415} & 0.412 & 0.434 & 0.427 & 0.445 \\
& Avg  & \textcolor{red}{0.329} & \textcolor{red}{0.374} & \textcolor{blue}{0.339} & \textcolor{blue}{0.380} & 0.364 & 0.395 & 0.383 & 0.407 \\
\hline
\multirow{5}{*}{ETTm1}
& 96  & \textcolor{red}{0.271} & \textcolor{red}{0.325} & \textcolor{blue}{0.310} & \textcolor{blue}{0.334} & 0.320 & 0.357 & 0.334 & 0.368 \\
& 192  & \textcolor{red}{0.322} & \textcolor{red}{0.351} & \textcolor{blue}{0.348} & \textcolor{blue}{0.362} & 0.361 & 0.381 & 0.390 & 0.393 \\
& 336  & \textcolor{red}{0.350} & \textcolor{red}{0.370} & \textcolor{blue}{0.376} & \textcolor{blue}{0.391} & 0.390 & 0.404 & 0.426 & 0.420 \\
& 720  & \textcolor{red}{0.414} & \textcolor{red}{0.404} & \textcolor{blue}{0.440} & \textcolor{blue}{0.423} & 0.454 & 0.441 & 0.491 & 0.459 \\
& Avg  & \textcolor{red}{0.339} & \textcolor{red}{0.362} & \textcolor{blue}{0.369} & \textcolor{blue}{0.378} & 0.381 & 0.395 & 0.407 & 0.410 \\
\hline
\multirow{5}{*}{ETTm2}
& 96  & \textcolor{red}{0.156} & \textcolor{red}{0.240} & \textcolor{blue}{0.170} & \textcolor{blue}{0.245} & 0.175 & 0.258 & 0.180 & 0.264 \\
& 192  & \textcolor{red}{0.212} & \textcolor{red}{0.280} & \textcolor{blue}{0.229} & \textcolor{blue}{0.291} & 0.237 & 0.299 & 0.250 & 0.309 \\
& 336  & \textcolor{red}{0.263} & \textcolor{red}{0.315} & 0.303 & 0.343 & \textcolor{blue}{0.298} & \textcolor{blue}{0.340} & 0.311 & 0.348 \\
& 720  & \textcolor{red}{0.351} & \textcolor{red}{0.371} & \textcolor{blue}{0.373} & 0.399 & 0.391 & \textcolor{blue}{0.396} & 0.412 & 0.407 \\
& Avg  & \textcolor{red}{0.245} & \textcolor{red}{0.301} & \textcolor{blue}{0.269} & \textcolor{blue}{0.320} & 0.275 & 0.323 & 0.288 & 0.332 \\
\hline
\multirow{5}{*}{Weather}
& 96  & \textcolor{red}{0.138} & \textcolor{red}{0.177} & \textcolor{blue}{0.155} & \textcolor{blue}{0.205} & 0.163 & 0.209 & 0.174 & 0.214 \\
& 192  & \textcolor{red}{0.181} & \textcolor{red}{0.220} & \textcolor{blue}{0.201} & \textcolor{blue}{0.245} & 0.208 & 0.250 & 0.221 & 0.254 \\
& 336  & \textcolor{red}{0.230} & \textcolor{red}{0.260} & \textcolor{blue}{0.237} & \textcolor{blue}{0.263} & 0.251 & 0.287 & 0.278 & 0.296 \\
& 720  & \textcolor{red}{0.304} & \textcolor{red}{0.315} & \textcolor{blue}{0.312} & \textcolor{blue}{0.334} & 0.339 & 0.341 & 0.358 & 0.347 \\
& Avg  & \textcolor{red}{0.213} & \textcolor{red}{0.243} & \textcolor{blue}{0.226} & \textcolor{blue}{0.262} & 0.240 & 0.271 & 0.258 & 0.278 \\
\hline
\multirow{5}{*}{Electricity}
& 96  & \textcolor{red}{0.126} & \textcolor{red}{0.217} & \textcolor{blue}{0.135} & \textcolor{blue}{0.222} & 0.153 & 0.247 & 0.148 & 0.240 \\
& 192  & \textcolor{red}{0.144} & \textcolor{red}{0.234} & \textcolor{blue}{0.147} & \textcolor{blue}{0.235} & 0.166 & 0.256 & 0.162 & 0.253 \\
& 336  & \textcolor{red}{0.158} & \textcolor{blue}{0.248} & \textcolor{blue}{0.164} & \textcolor{red}{0.245} & 0.185 & 0.277 & 0.178 & 0.269 \\
& 720  & \textcolor{red}{0.190} & \textcolor{red}{0.277} & \textcolor{blue}{0.212} & \textcolor{blue}{0.310} & 0.225 & \textcolor{blue}{0.310} & 0.225 & 0.317 \\
& Avg  & \textcolor{red}{0.154} & \textcolor{red}{0.244} & \textcolor{blue}{0.165} & \textcolor{blue}{0.253} & 0.182 & 0.272 & 0.178 & 0.270 \\
\hline
\multirow{5}{*}{Traffic}
& 96  & \textcolor{red}{0.343} & \textcolor{red}{0.225} & \textcolor{blue}{0.392} & \textcolor{blue}{0.253} & 0.462 & 0.285 & 0.395 & 0.268 \\
& 192  & \textcolor{red}{0.369} & \textcolor{red}{0.238} & \textcolor{blue}{0.402} & \textcolor{blue}{0.258} & 0.473 & 0.296 & 0.417 & 0.276 \\
& 336  & \textcolor{red}{0.382} & \textcolor{red}{0.242} & \textcolor{blue}{0.428} & \textcolor{blue}{0.263} & 0.498 & 0.296 & 0.433 & 0.283 \\
& 720  & \textcolor{red}{0.424} & \textcolor{red}{0.270} & \textcolor{blue}{0.441} & \textcolor{blue}{0.282} & 0.506 & 0.313 & 0.467 & 0.302 \\
& Avg  & \textcolor{red}{0.379} & \textcolor{red}{0.243} & \textcolor{blue}{0.416} & \textcolor{blue}{0.264} & 0.484 & 0.297 & 0.428 & 0.282 \\
\hline
\end{tabular}
} 
\caption{Comparison of EMTSF (Ours), TimeMixer++, TimeMixer (2024b), and iTransformer (2024) across multiple datasets and horizons using MSE and MAE metrics.}
\label{tab:ComparisonTimeMixeriTrans}
\end{table}

\begin{table}[h] 
\centering
\scalebox{0.65}{ 
\scriptsize 
\setlength{\tabcolsep}{2pt} 
\begin{tabular}{|c|c|cc|cc|cc|cc|cc|cc|cc|}
\hline
\textbf{Dataset} & \textbf{Horizon}
& \multicolumn{2}{c|}{\textbf{EMTSF (Ours)}}
& \multicolumn{2}{c|}{\textbf{TIME-MOE$_\text{base}$}}
& \multicolumn{2}{c|}{\textbf{TIME-MOE$_\text{large}$}}
& \multicolumn{2}{c|}{\textbf{TIME-MOE$_\text{ultra}$}}
& \multicolumn{2}{c|}{\textbf{Moirai$_\text{small}$}}
& \multicolumn{2}{c|}{\textbf{Moirai$_\text{base}$}}
& \multicolumn{2}{c|}{\textbf{Moirai$_\text{large}$}} \\
\cline{3-16}
& & \textbf{MSE} & \textbf{MAE}
& \textbf{MSE} & \textbf{MAE}
& \textbf{MSE} & \textbf{MAE}
& \textbf{MSE} & \textbf{MAE}
& \textbf{MSE} & \textbf{MAE}
& \textbf{MSE} & \textbf{MAE}
& \textbf{MSE} & \textbf{MAE} \\
\hline
& 96 & 0.359 & 0.384 & 0.357 & \textcolor{blue}{0.381} & \textcolor{blue}{0.350} & 0.382 & \textcolor{red}{0.349} & \textcolor{red}{0.379} & 0.401 & 0.402 & 0.376 & 0.392 & 0.381 & 0.388 \\
 & 192 & 0.399 & 0.411 & \textcolor{red}{0.384} & \textcolor{blue}{0.404} & \textcolor{blue}{0.388} & 0.412 & 0.395 & 0.413 & 0.435 & 0.421 & 0.412 & 0.413 & 0.431 & \textcolor{red}{0.400} \\
ETTh1 & 336 &  \textcolor{blue}{0.418} & \textcolor{red}{0.422} & \textcolor{red}{0.411} & 0.434 & \textcolor{red}{0.411} & 0.430 & 0.447 & 0.453 & 0.453 & 0.433 & 0.443 & 0.425 & 0.495 & 0.430 \\
 & 720 & \textcolor{blue}{0.436} & \textcolor{blue}{0.454} & 0.449 & 0.477 & \textcolor{red}{0.427} & 0.455 & 0.457 & 0.462 & 0.439 & 0.454 & 0.487 & \textcolor{red}{0.444} & 0.611 & 0.456 \\
 & Avg & 0.403 & \textcolor{red}{0.417} & \textcolor{blue}{0.400} &
 0.424 & \textcolor{red}{0.394} & \textcolor{blue}{0.419} & 0.412 & 0.426 & 0.428 & 0.427 & 0.417 &\textcolor{blue}{ 0.419} & 0.480 & \textcolor{blue}{ 0.419} \\
 \hline
 & 96 & \textcolor{red}{0.262} & 0.354 & 0.305 & 0.359 & 0.302 & 0.354 & \textcolor{blue}{0.292} & 0.352 & 0.297 & \textcolor{blue}{ 0.336 }& 0.294 & \textcolor{red}{0.330} & 0.296 & \textcolor{red}{0.330} \\
 & 192 & \textcolor{red}{0.328} & \textcolor{red}{0.371} & 0.351 & 0.386 & 0.364 & 0.386 & \textcolor{blue}{0.347} & 0.379 & 0.362 & \textcolor{blue}{0.375} & 0.363 & \textcolor{red}{0.371} & 0.370 & \textcolor{red}{0.371} \\
ETTh2 & 336 & \textcolor{red}{0.347} & \textcolor{blue}{0.387} & 0.391 & 0.418 & 0.417 & 0.425 & \textcolor{blue}{0.366} & 0.419 & 0.370 & 0.393 & 0.376 & 0.390 & 0.393 & \textcolor{red}{0.384} \\
 & 720 & \textcolor{red}{0.381} & \textcolor{red}{0.417} & 0.419 & 0.454 & 0.537 & 0.496 & 0.439 & 0.447 & \textcolor{blue}{0.411} & 0.426 & 0.416 & 0.433 & 0.423 & \textcolor{blue}{0.418} \\
 & Avg & \textcolor{red}{0.329} & \textcolor{red}{0.374} & 0.366 & 0.404 & 0.405 & 0.415 & 0.371 & 0.399 & \textcolor{blue}{0.361} & 0.384 & 0.362 & 0.382 & 0.367 & \textcolor{blue}{0.377} \\
 \hline
 & 96 & \textcolor{red}{0.271} & \textcolor{red}{0.325} & 0.338 & 0.368 & 0.309 & 0.357 & \textcolor{blue}{0.281} & \textcolor{blue}{0.341} & 0.418 & 0.392 & 0.363 & 0.356 & 0.380 & 0.361 \\
 & 192 & \textcolor{blue}{0.322} & \textcolor{red}{0.351} & 0.353 & 0.388 & 0.346 & 0.381 & \textcolor{red}{0.305} & \textcolor{blue}{0.358} & 0.431 & 0.405 & 0.388 & 0.375 & 0.412 & 0.388 \\
ETTm1 & 336 & \textcolor{red}{0.350} & \textcolor{red}{0.370} & 0.381 & 0.413 & 0.373 & 0.408 & \textcolor{blue}{0.360} & \textcolor{blue}{0.395} & 0.460 & 0.418 & 0.392 & 0.436 & 0.434 & 0.404 \\
 & 720 & \textcolor{red}{0.414} & \textcolor{red}{0.404} & 0.504 & 0.493 & 0.475 & 0.477 & 0.469 & 0.472 & 0.462 & 0.432 & \textcolor{blue}{0.460} & \textcolor{blue}{0.418} & 0.462 & 0.420 \\
 & Avg & \textcolor{red}{0.339} & \textcolor{red}{0.362} & 0.394 & 0.415 & 0.376 & 0.405 & \textcolor{blue}{0.356} & 0.391 & 0.436 & 0.410 & 0.406 & \textcolor{blue}{0.385} & 0.422 & 0.391 \\
 \hline
 & 96 & \textcolor{red}{0.156} & \textcolor{red}{0.240} & 0.201 & 0.291 & \textcolor{blue}{0.197} & 0.286 & 0.198 & 0.288 & 0.214 & 0.288 & 0.205 & \textcolor{blue}{0.273} & 0.211 & 0.274 \\
 & 192 & \textcolor{red}{0.212} & \textcolor{red}{0.280} & 0.258 & 0.334 & 0.250 & 0.322 & \textcolor{blue}{0.235} & \textcolor{blue}{0.312} & 0.284 & 0.332 & 0.275 & 0.316 & 0.281 & 0.318 \\
ETTm2 & 336 & \textcolor{red}{0.263} & \textcolor{red}{0.315} & 0.324 & 0.373 & 0.337 & 0.375 & \textcolor{blue}{0.293} & \textcolor{blue}{0.348} & 0.331 & 0.362 & 0.329 & 0.350 & 0.341 & 0.355 \\
 & 720 & \textcolor{red}{0.351} & \textcolor{red}{0.371} & 0.488 & 0.464 & 0.480 & 0.461 & 0.427 & 0.428 & 0.402 & \textcolor{blue}{0.408} & \textcolor{blue}{0.385} & 0.423 & 0.485 & 0.426 \\
 & Avg & \textcolor{red}{0.245} & \textcolor{red}{0.301} & 0.317 & 0.365 & 0.316 & 0.361 & \textcolor{blue}{0.288} & 0.344 & 0.307 & 0.347 & 0.337 & \textcolor{blue}{0.337} & 0.329 & 0.343 \\
 \hline
 & 96 & \textcolor{red}{0.138} & \textcolor{red}{0.177} & 0.160 & 0.214 & 0.159 & 0.213 & \textcolor{blue}{0.157} & \textcolor{blue}{0.211} & 0.198 & 0.222 & 0.220 & 0.217 & 0.199 &\textcolor{blue}{ 0.211} \\
 & 192 & \textcolor{red}{0.181} & \textcolor{red}{0.220} & 0.210 & 0.260 & 0.215 & 0.266 & \textcolor{blue}{0.208} & 0.256 & 0.247 & 0.265 & 0.271 & 0.259 & 0.246 & \textcolor{blue}{0.251} \\
Weather & 336 & \textcolor{red}{0.230} & \textcolor{red}{0.260} & 0.274 & 0.309 & 0.291 & 0.322 & \textcolor{blue}{0.255} & \textcolor{blue}{0.290} & 0.283 & 0.303 & 0.286 & 0.297 & 0.274 & 0.291 \\
 & 720 & \textcolor{red}{0.304} & \textcolor{red}{0.315} & 0.418 & 0.405 & 0.415 & 0.400 & 0.405 & 0.397 & 0.373 & 0.354 & 0.373 & 0.354 & \textcolor{blue}{0.337} & \textcolor{blue}{0.340} \\
 & Avg & \textcolor{red}{0.213} & \textcolor{red}{0.243} & 0.265 & 0.297 & 0.270 & 0.300 & \textcolor{blue}{0.256} & 0.288 & 0.275 & 0.286 & 0.287 & 0.281 & 0.264 & \textcolor{blue}{0.273} \\
\hline
\end{tabular}
} 
\caption{Comparison of EMTSF (ours) with TIME-MOE and MOIRAI-MOE using MSE and MAE (Red = Best, Blue = 2nd Best)}
\label{tab:ComparisonMOIRAITimeMOE}
\end{table}

\begin{table}[h] 
\centering
\scalebox{0.72}{ 
\scriptsize 
\setlength{\tabcolsep}{2pt} 
\begin{tabular}{|c|c|c|c|c|c|c|}
\hline
\textbf{Models}    & \textbf{Horizon} & \textbf{EMTSF (MoE)} & \textbf{xLSTMTime} & \textbf{PatchTST} & \textbf{minGRUTime} & \textbf{ELM} \\ \hline
\textbf{Metric}    & \textbf{}        & \textbf{MSE\hspace{0.5cm}MAE} & \textbf{MSE\hspace{0.5cm}MAE} & \textbf{MSE\hspace{0.5cm}MAE} & \textbf{MSE\hspace{0.5cm}MAE} & \textbf{MSE\hspace{0.5cm}MAE} \\ \hline
\textbf{Weather}   & 96 & \textcolor{red}{0.138}\hspace{0.5cm}\textcolor{red}{0.177} & 0.147\hspace{0.5cm}0.192 & \textcolor{blue}{0.143}\hspace{0.5cm}\textcolor{blue}{0.180} & \textcolor{blue}{0.143}\hspace{0.5cm}0.182 & 0.160\hspace{0.5cm}0.187 \\ 
                   & 192 & \textcolor{red}{0.182}\hspace{0.5cm}\textcolor{red}{0.220} & 0.194\hspace{0.5cm}0.237 & 0.187\hspace{0.5cm}0.224 & 0.187\hspace{0.5cm}0.228 & \textcolor{blue}{0.185}\hspace{0.5cm}\textcolor{blue}{0.222} \\ 
                   & 336 & \textcolor{red}{0.232}\hspace{0.5cm}\textcolor{red}{0.260} & 0.240\hspace{0.5cm}0.275 & 0.240\hspace{0.5cm}0.267 & \textcolor{blue}{0.236}\hspace{0.5cm}0.266 & 0.242\hspace{0.5cm}\textcolor{blue}{0.264} \\  
                   & 720 & \textcolor{red}{0.305}\hspace{0.5cm}\textcolor{red}{0.315} & 0.315\hspace{0.5cm}0.329 & 0.312\hspace{0.5cm}\textcolor{blue}{0.318} & \textcolor{blue}{0.311}\hspace{0.5cm}0.320 & \textcolor{blue}{0.311}\hspace{0.5cm}0.319 \\ \hline
\textbf{Traffic}   & 96 &  \textcolor{blue}{ 0.343} \hspace{0.5cm}\textcolor{red}{0.225}    & \textcolor{red}{0.341} \hspace{0.5cm} \textcolor{blue}{0.229} &0.376\hspace{0.5cm}	0.241 & 0.396\hspace{0.5cm} 0.265 & 0.412\hspace{0.5cm}0.263 \\  
                   & 192   & \textcolor{blue}{0.369}\hspace{0.5cm}\textcolor{red}{0.238} &\textcolor{red}{0.365}\hspace{0.5cm}0.242 & 0.387\hspace{0.5cm}\textcolor{blue}{0.241} & 0.415\hspace{0.5cm}0.269 & 0.425\hspace{0.5cm}0.266 \\  
                   & 336  & \textcolor{red}{0.382}\hspace{0.5cm}\textcolor{red}{0.242} & \textcolor{red}{0.382}\hspace{0.5cm}\textcolor{blue}{0.245} & \textcolor{blue}{0.398}\hspace{0.5cm}0.246 & 0.424\hspace{0.5cm}0.275 & 0.431\hspace{0.5cm}0.270 \\  
                   & 720  & \textcolor{red}{0.424}\hspace{0.5cm}\textcolor{red}{0.270} & \textcolor{blue}{0.425}\hspace{0.5cm} \textcolor{red}{0.270} & 0.437\hspace{0.5cm}\textcolor{blue}{0.272} & 0.458\hspace{0.5cm}0.300 & 0.465\hspace{0.5cm}0.289 \\ \hline

\textbf{Electricity} & 96  & \textcolor{red}{0.126\hspace{0.5cm}0.217} & 0.130\hspace{0.5cm}0.223 & \textcolor{blue}{0.127}\hspace{0.5cm}\textcolor{blue}{0.218} & 0.131\hspace{0.5cm}0.223  &0.132\hspace{0.5cm}0.223\\   
                   & 192  &\textcolor{red}{0.144}\hspace{0.5cm}\textcolor{blue}{0.234} & 0.149\hspace{0.5cm}0.241 & \textcolor{red}{0.144}\hspace{0.5cm}\textcolor{red}{0.233} & 0.149\hspace{0.5cm}0.240         &\textcolor{blue}{0.147}\hspace{0.5cm}0.236                \\  
                   & 336  &\textcolor{red}{ 0.158\hspace{0.5cm}0.248} & 0.165\hspace{0.5cm}0.258 & \textcolor{blue}{0.160}\hspace{0.5cm}\textcolor{blue}{0.250} & 0.166\hspace{0.5cm}0.257 & 0.163\hspace{0.5cm}0.253 \\  
                   & 720  & \textcolor{red}{ 0.190\hspace{0.5cm}0.277} & \textcolor{blue}{0.191}\hspace{0.5cm}\textcolor{blue}{0.279} & 0.193\hspace{0.5cm}\textcolor{blue}{0.279} & 0.205\hspace{0.5cm}0.291 & 0.204\hspace{0.5cm}0.288 \\ \hline
\textbf{Illness}   & 24   & \textcolor{red}{ 1.617\hspace{0.5cm}0.732} & \textcolor{blue}{1.639}\hspace{0.5cm}\textcolor{blue}{0.733} & 1.670\hspace{0.5cm}0.784 & 1.665\hspace{0.5cm}0.750 & 1.939\hspace{0.5cm}0.808 \\  
                   & 36   & \textcolor{blue} {1.586}\hspace{0.5cm} \textcolor{blue} {0.728} & \textcolor{red}{1.486}\hspace{0.5cm} \textcolor{red}{0.711 }& 1.709\hspace{0.5cm}0.789 & 1.648\hspace{0.5cm}0.758 & 1.838\hspace{0.5cm}0.806 \\ 
                   & 48   & \textcolor{blue}{1.587}\hspace{0.5cm}\textcolor{blue}{0.753} & \textcolor{red}{1.461} \hspace{0.5cm} \textcolor{red}{0.742} & 1.748\hspace{0.5cm}0.800 & 1.599\hspace{0.5cm}0.762 & 1.783\hspace{0.5cm}0.817 \\  
                   & 60  & \textcolor{blue}{1.560} \hspace{0.5cm} \textcolor{red}{0.768} & 1.617\hspace{0.5cm}0.798 & 1.608\hspace{0.5cm}0.790 & \textcolor{red}{1.543}\hspace{0.5cm}\textcolor{blue}{0.773} & 1.760\hspace{0.5cm}0.836 \\ \hline
\textbf{ETTh1}     & 96   & \textcolor{red}{0.359}\hspace{0.5cm}\textcolor{red}{0.384} & 0.444\hspace{0.5cm}0.455 & 0.372\hspace{0.5cm}0.397 & \textcolor{blue}{0.362}\hspace{0.5cm}\textcolor{blue}{0.387} & 0.372\hspace{0.5cm}0.389 \\   
                   & 192  & \textcolor{red}{ 0.399\hspace{0.5cm}0.411} & 0.463\hspace{0.5cm}0.462 & 0.429\hspace{0.5cm}0.434 & \textcolor{blue}{0.402\hspace{0.5cm}0.414 }& 0.412\hspace{0.5cm}0.416 \\  
                   & 336  & \textcolor{red}{0.418\hspace{0.5cm}0.422} & 0.487\hspace{0.5cm}0.481 & 0.438\hspace{0.5cm}0.438 & \textcolor{blue}{0.436\hspace{0.5cm}0.429} & 0.448\hspace{0.5cm}0.433 \\  
                   & 720  & \textcolor{red}{0.436}\hspace{0.5cm}\textcolor{red}{0.454 }& 0.526\hspace{0.5cm}0.523 & 0.460\hspace{0.5cm}0.468 & \textcolor{blue}{0.446\hspace{0.5cm}0.455} & 0.476\hspace{0.5cm}0.461 \\ \hline
\textbf{ETTh2}     & 96   & \textcolor{red}{0.262\hspace{0.5cm}0.324} & 0.281\hspace{0.5cm}0.342 & 0.279\hspace{0.5cm}0.336 & 0.272\hspace{0.5cm}\textcolor{blue}{0.331} &\textcolor{blue}{0.269\hspace{0.5cm}0.331} \\  
                   & 192  & \textcolor{red}{0.328}\hspace{0.5cm}\textcolor{red}{0.371} & 0.357\hspace{0.5cm}0.387 & 0.352\hspace{0.5cm}0.385 & 0.341\hspace{0.5cm}0.377 &\textcolor{blue}{ 0.331\hspace{0.5cm}0.372} \\  
                   & 336  & \textcolor{red}{ 0.347\hspace{0.5cm}0.387} & 0.397\hspace{0.5cm}0.421 & 0.367\hspace{0.5cm}0.400 & 0.363\hspace{0.5cm}0.400 &\textcolor{blue}{ 0.358\hspace{0.5cm}0.399} \\ 
                   & 720  & \textcolor{red}{0.381}\hspace{0.5cm}\textcolor{red}{0.417} & 0.410\hspace{0.5cm}0.438 &\textcolor{blue} {0.399}\hspace{0.5cm}0.430 & 0.405\hspace{0.5cm}0.432 & 0.438\hspace{0.5cm}\textcolor{blue}{0.424} \\ \hline
\textbf{ETTm1}     & 96   &\textcolor{red}{ 0.271}\hspace{0.5cm}\textcolor{red}{0.325} & 0.288\hspace{0.5cm}0.335 & \textcolor{blue}{0.284}\hspace{0.5cm}0.332 & 0.288\hspace{0.5cm}0.334 & 0.294\hspace{0.5cm}\textcolor{blue}{0.331} \\ 
                   & 192  & \textcolor{red}{0.322\hspace{0.5cm}0.351} & 0.330\hspace{0.5cm}0.364 & 0.335\hspace{0.5cm}0.361 & \textcolor{blue}{0.328}\hspace{0.5cm}0.358 & 0.337\hspace{0.5cm}\textcolor{blue}{0.356} \\  
                   & 336  & \textcolor{red}{0.350\hspace{0.5cm}0.370} & \textcolor{blue}{0.363}\hspace{0.5cm}0.383 & 0.365\hspace{0.5cm}0.380 & 0.364\hspace{0.5cm}0.379 & 0.370\hspace{0.5cm}\textcolor{blue}{0.376} \\  
                   & 720  & \textcolor{red}{0.414}\hspace{0.5cm}\textcolor{red}{0.404} & \textcolor{blue}{0.419}\hspace{0.5cm}0.414 & 0.420\hspace{0.5cm}0.410 & 0.424\hspace{0.5cm}0.411 & 0.426\hspace{0.5cm}\textcolor{blue}{0.408 }\\ \hline
\textbf{ETTm2}     & 96   &\textcolor{red}{ 0.156\hspace{0.5cm}0.240} & 0.166\hspace{0.5cm}0.253 & 0.160\hspace{0.5cm}0.246 & 0.160\hspace{0.5cm}0.246 &\textcolor{blue} {0.159\hspace{0.5cm}0.243} \\ 
                   & 192  & \textcolor{red}{0.212}\hspace{0.5cm}\textcolor{red}{0.280} & 0.224\hspace{0.5cm}0.293 & 0.246\hspace{0.5cm}0.290 & \textcolor{blue}{0.216}\hspace{0.5cm}0.286 &\textcolor{blue}{ 0.216}\hspace{0.5cm}\textcolor{blue}{0.282} \\  
                   & 336  & \textcolor{red}{0.263\hspace{0.5cm}0.315} & 0.278\hspace{0.5cm}0.329 &\textcolor{blue}{ 0.265\hspace{0.5cm}0.317} & 0.267\hspace{0.5cm}0.319 & 0.272\hspace{0.5cm}0.320 \\  
                   & 720  & \textcolor{blue}{0.351}\hspace{0.5cm}\textcolor{red}{0.371} & 0.367\hspace{0.5cm}0.385 &\textcolor{red}{ 0.350}\hspace{0.5cm}\textcolor{blue}{0.372} & 0.358\hspace{0.5cm}0.376 & 0.360\hspace{0.5cm}0.378 \\ \hline
\end{tabular}%
}
\caption{Performance Comparison of EMTSF MoE Design with Component Models on popular TSF Datasets}
\label{tab:ComparisonComponentModels}
\end{table}

\subsection{Ablation Study - 
 EMTSF MoE Design}
Our EMTSF design initially uses four complementary models in the MoE framework. The models used are: PatchTST, xLSTMTime, minGRUTime and ELM (Enhanced Linear Model). Even though these models perform well in TSF, do they effectively combine in an MoE framework to produce better results than individual models? To answer this question, we separately trained each of the models in EMTSF and compared their results with the EMTSF end-to-end MoE trained model (architecture described in Figure \ref{fig:FigureMoE}).
Table \ref{tab:ComparisonComponentModels} shows the comparisons of our EMTSF MoE with the individual models in it on the popular datasets for TSF. The forecasting is done for prediction targets of $T$ =  \{96, 192, 336, 720\} with a look-back window of $L$= 512. This table shows that EMTSF is significantly better than any of the component models (i.e., xLSTMTime, PatchTST, minGRUTime and ELM). Further, the second best model varies depending upon the dataset and the length of prediction indicating that the contribution from the expert models is data dependent, and all models complement each other in this respect. As can be seen, the mixture of experts is combining the individual experts in a cooperative manner to enhance the overall accuracy in prediction on the various datasets.

\begin{figure}
    \centering
    \includegraphics[width=1.0\linewidth]{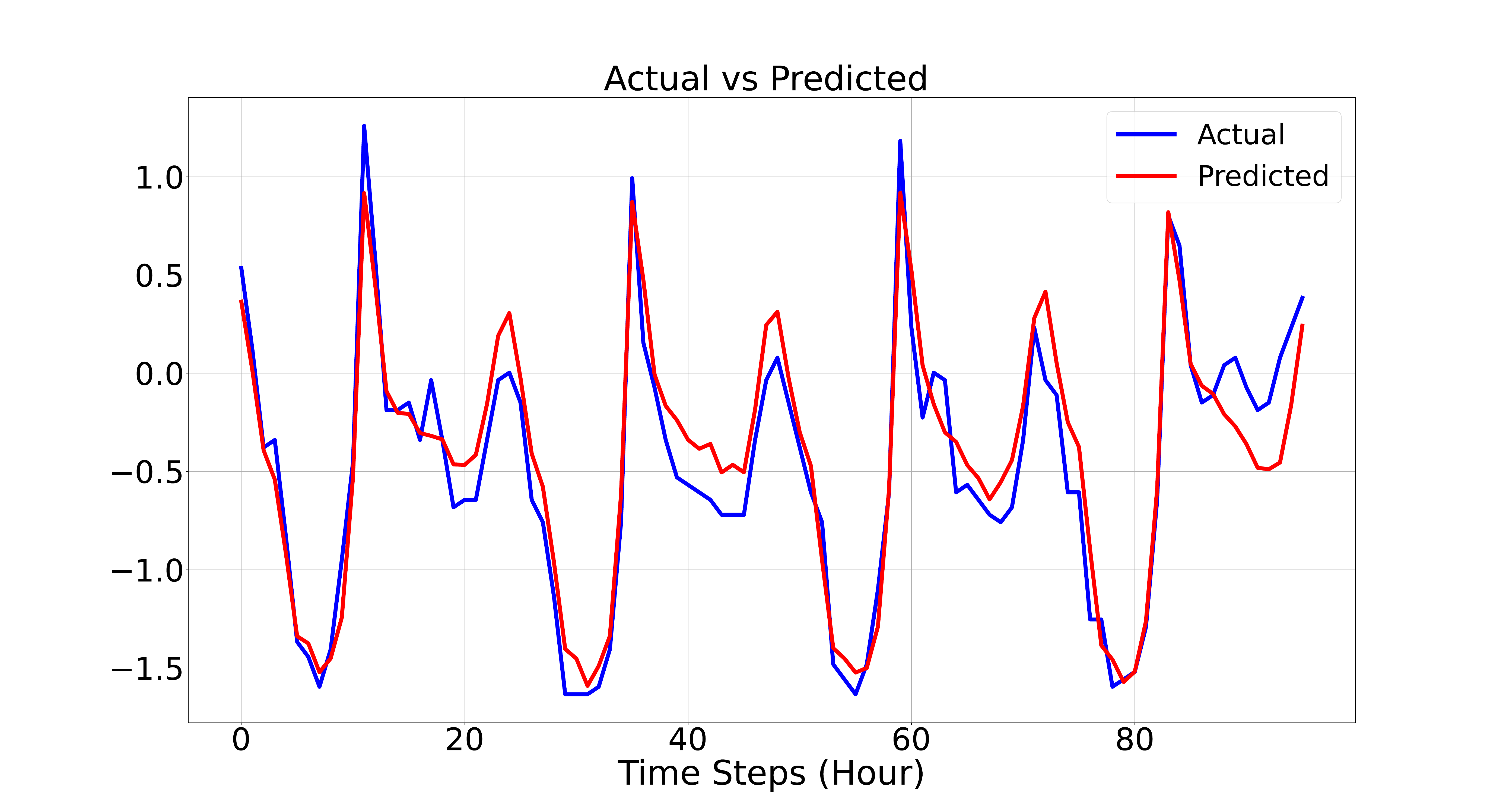}
    \caption{Actual vs. Predicted for the Electricity Dataset}
    \label{fig:electricityactualvspredicted}
\end{figure}

\begin{figure}
    \centering
    \includegraphics[width=1.0\linewidth]{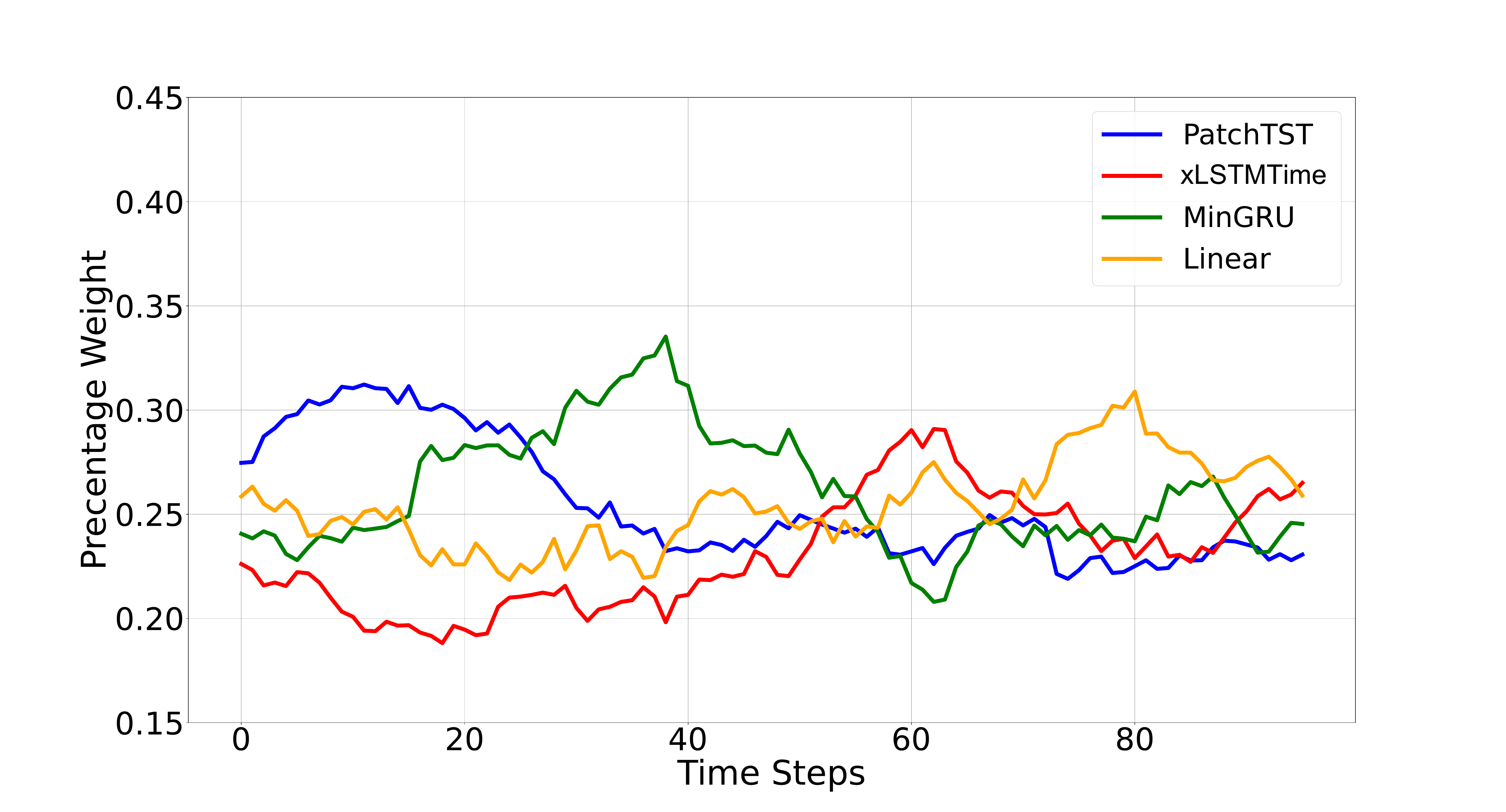}
\caption{Percentage Gating Weights of different Models in EMTSF for Electricity Dataset}
\label{fig:ElectricityGatingWeight}
\end{figure}

\begin{figure}
    \centering
    \includegraphics[width=1.0\linewidth]{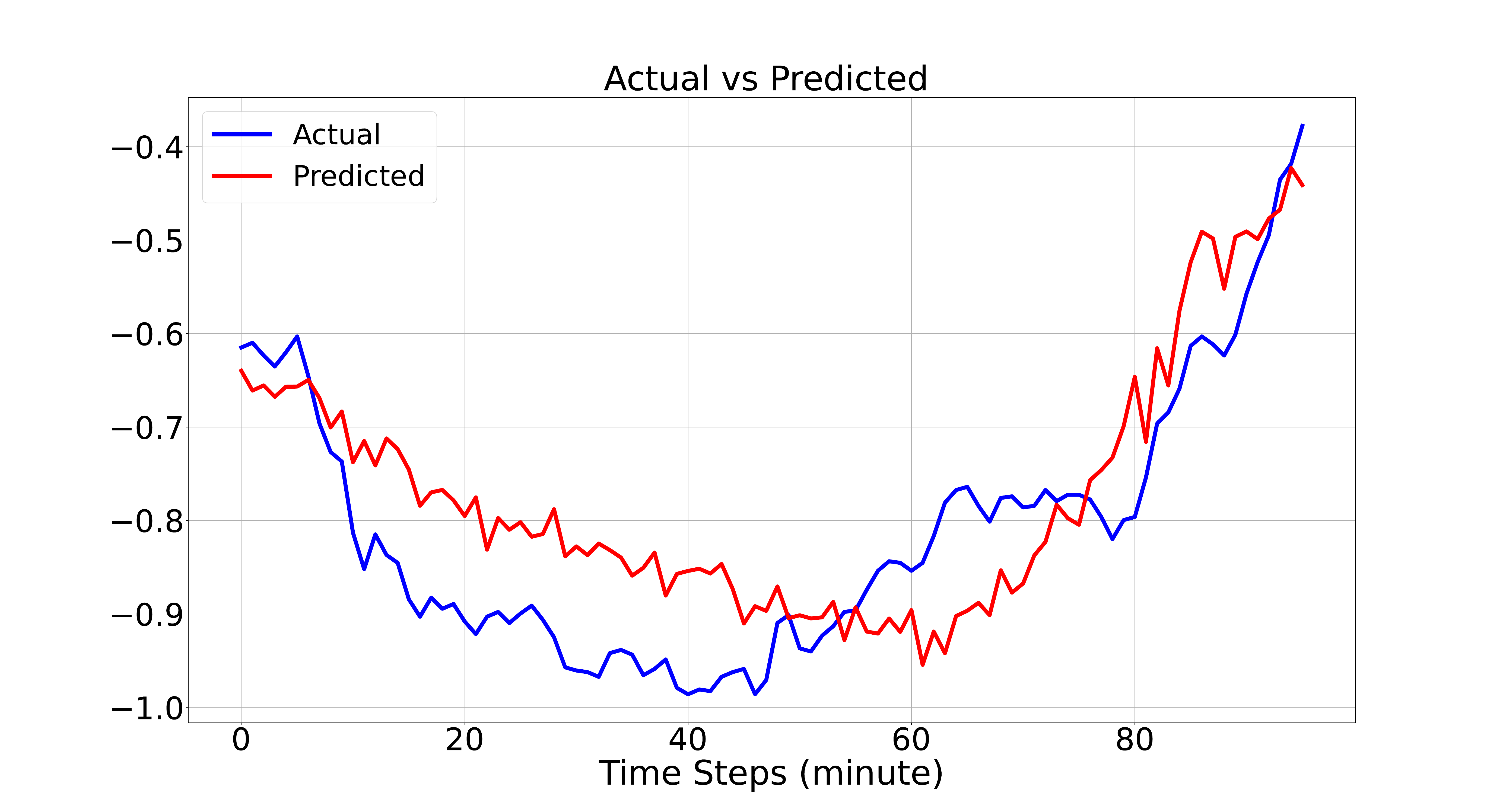}
    \caption{Actual vs. Predicted for the Weather Dataset}
    \label{fig:weatheractualvspredicted}
\end{figure}

\begin{figure}
    \centering
    \includegraphics[width=1.0\linewidth]{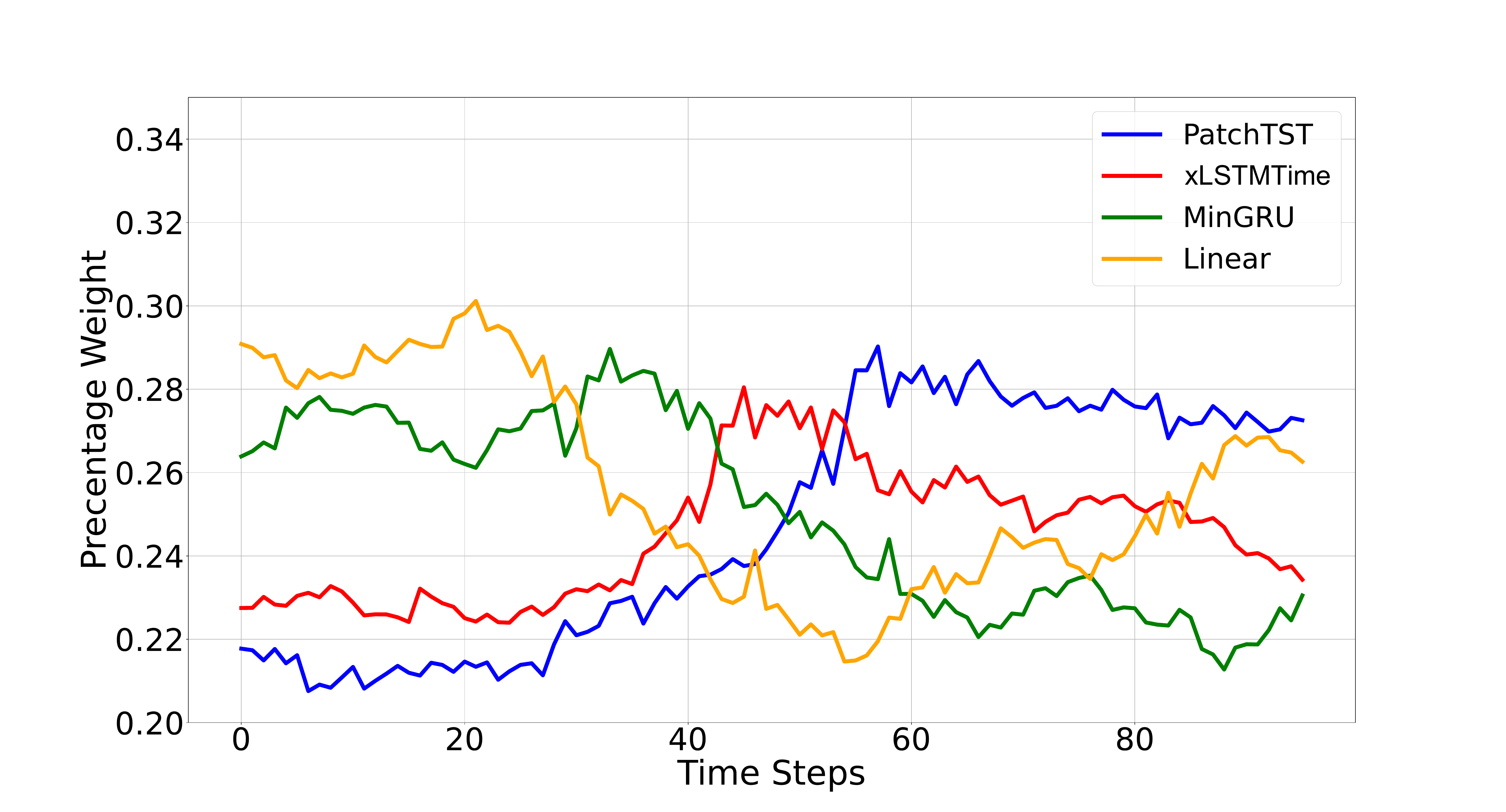}
    \caption{Percentage Gating Weights of different Models in EMTSF for Weather Dataset}
    \label{fig:WeatherGatingWeight}
\end{figure}

\begin{figure}
    \centering
 \includegraphics[width=1.0\linewidth]{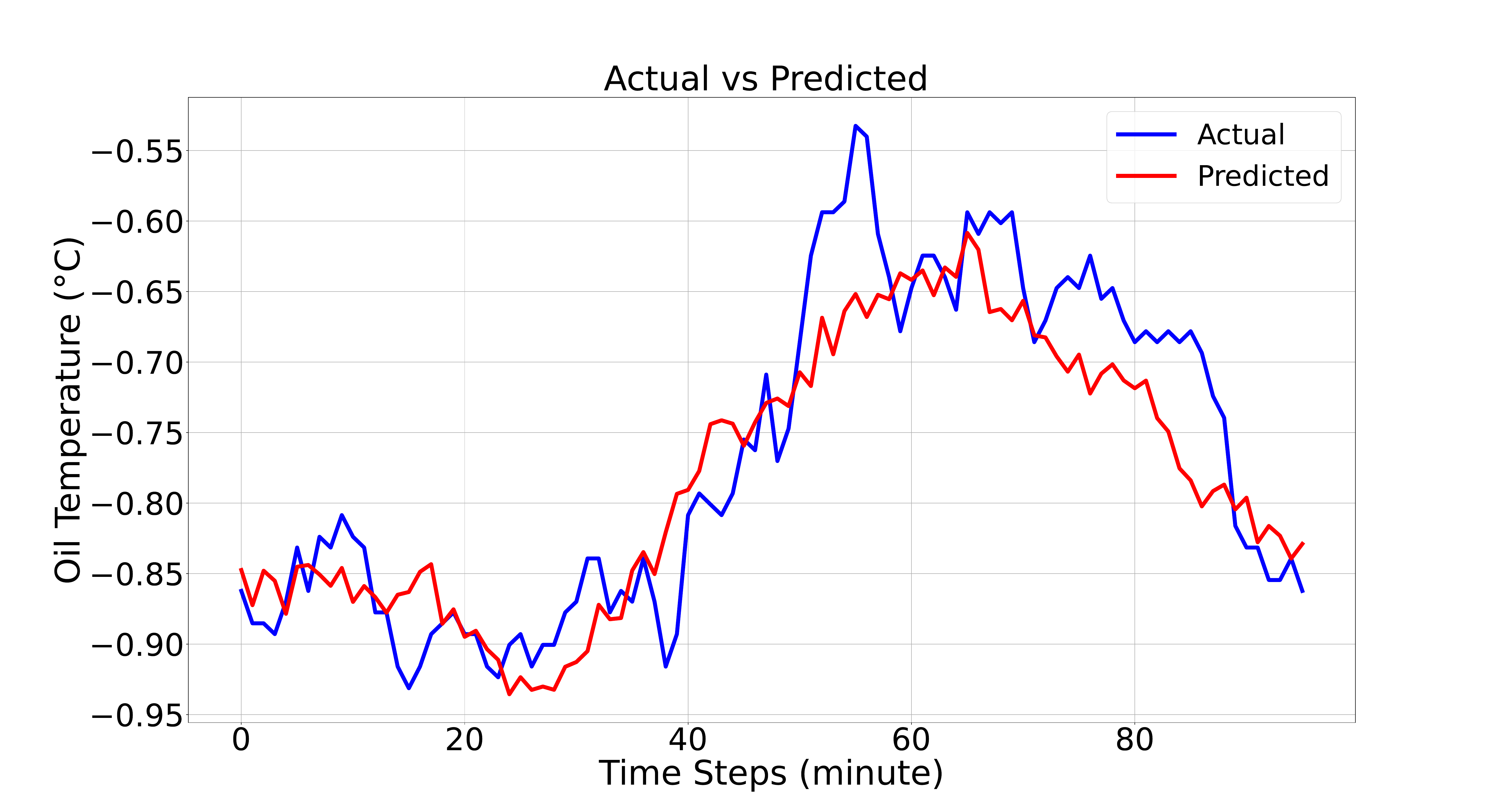}
    \caption{Actual vs. Predicted for the Ettm1 Dataset}
    \label{fig:Ettm1ActualvsPredicted}
\end{figure}

\begin{figure}
    \centering
\includegraphics[width=1.0\linewidth]{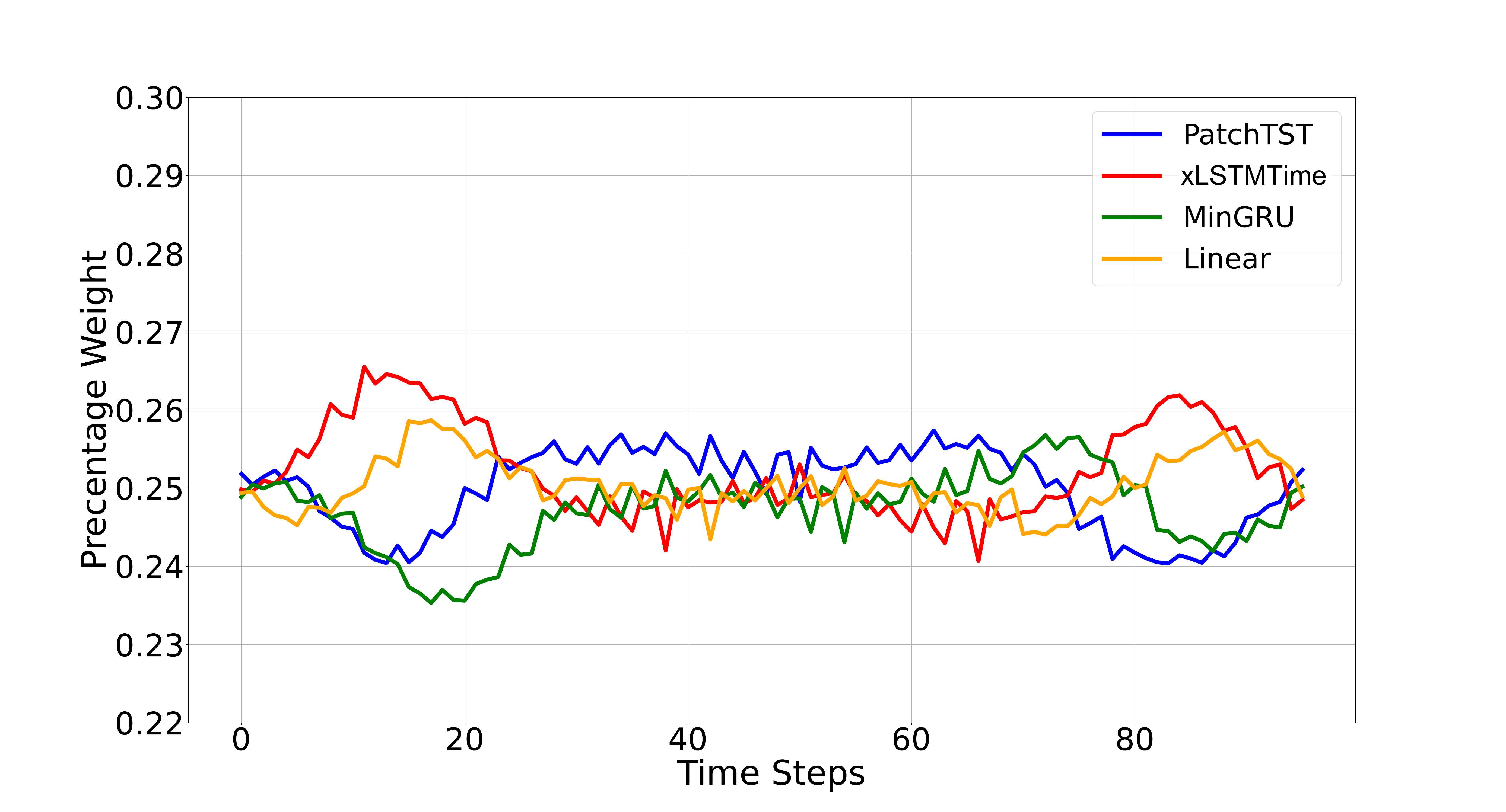}
    \caption{Percentage Gating Weights of different Models in EMTSF for Ettm1 Dataset}
\label{fig:Ettm1GatingWeights}
\end{figure}

Figure \ref{fig:electricityactualvspredicted} shows the graphs for actual versus predicted time series values for the electricity dataset. Our model learns the periodicity and the variations in the data very nicely. To determine, how well the different experts in our EMTSF MoE are contributing to overall generated output,
Figure \ref{fig:ElectricityGatingWeight} 
 shows the gating weights for each time step in the prediction for the electricity dataset. The Transformer-based EMTSF gating network selects the weight of each expert according to the learnt contribution for that expert. As can be seen, each expert's weight varies temporally according to the dynamics of the data. Figure \ref{fig:weatheractualvspredicted} shows the prediction for the weather dataset by our model. Figure \ref{fig:WeatherGatingWeight} shows the contribution of each expert for the weather dataset. Similarly, Figures \ref{fig:Ettm1ActualvsPredicted} and \ref{fig:Ettm1GatingWeights} show the same for the Ettm1 dataset. 

\begin{figure}
    \centering
    \includegraphics[width=1.0\linewidth]{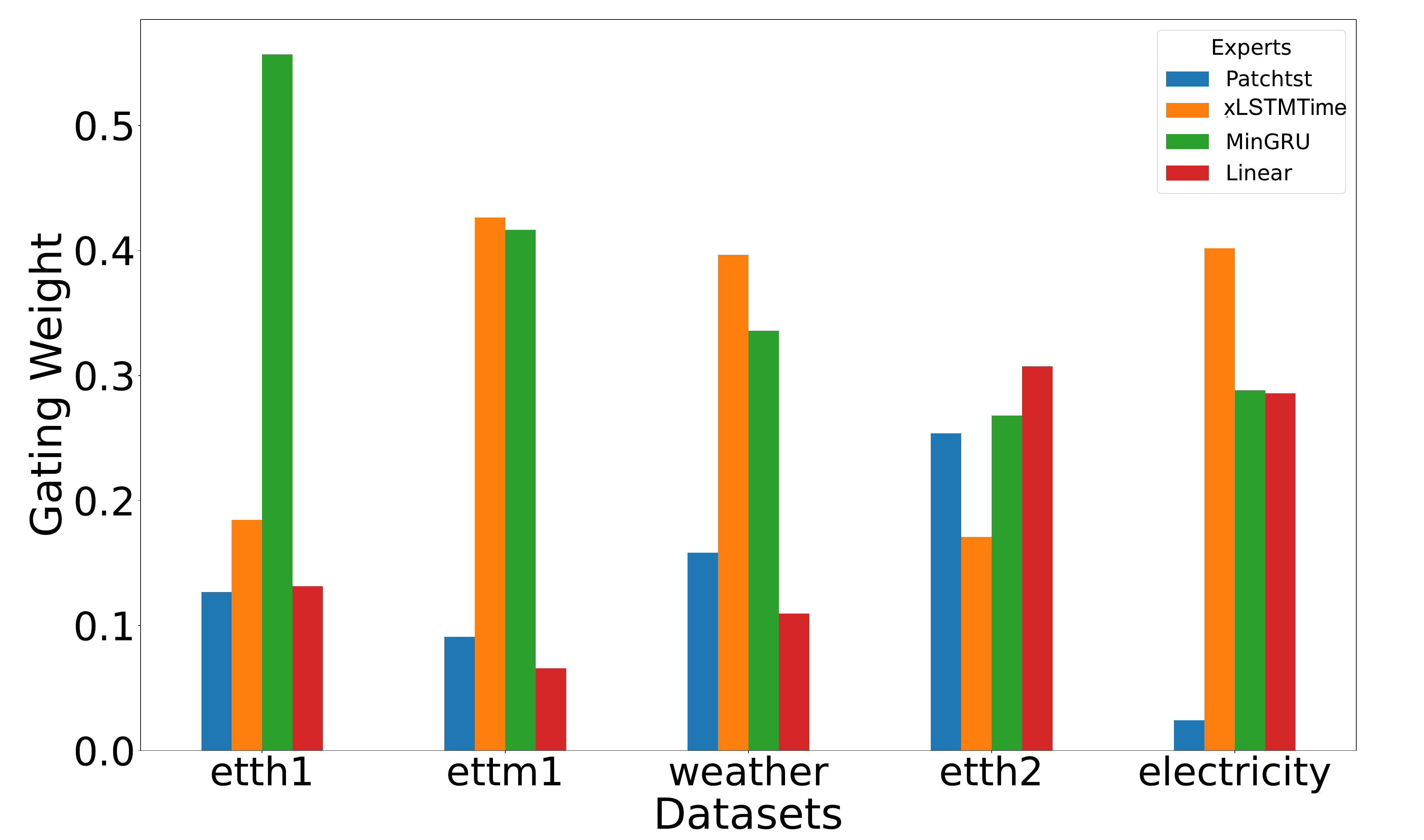}
    \caption{Percentage Gating Weights for different Models in EMTSF for different Datasets}
    \label{fig:BarGraphGatingWeights}
\end{figure}

Figure \ref{fig:BarGraphGatingWeights} shows the average weight over the predicted length for each expert assigned by the Gating Network during the output generation for different datasets. All four experts provide a significant contribution to the overall generated output. This further confirms that there is no expert collapse in our MoE design.

\vspace{0.2cm}
Table \ref{tab:PEMSComparison1} shows comparisons of our EMTSF model (MoE) with the individual models in EMTSF on the PEMS datasets. The forecasting is done with prediction targets of $T$ = \{12, 24, 48, 96\} while using a look-back window of $L$= 96. It can be seen from  Table \ref{tab:PEMSComparison1} that for this dataset also, the EMTSF is  better than any of the component models (i.e., xLSTMTime, PatchTST, minGRUTime and ELM) individually. Even though, for this dataset xLSTMTime performs better than other component models, the MoE design in EMTSF that uses all the models in a cooperative and complementary manner is significantly more effective. Average performance gain by EMTSF over the second best model is also indicated in Table \ref{tab:PEMSComparison1}.

\begin{table}[h]
\centering
\scalebox{0.75}{
\scriptsize
\setlength{\tabcolsep}{2pt}
\begin{tabular}{|c|c|c|c|c|c|c|}
\hline
\textbf{Models} & \textbf{Horizon} & \textbf{EMTSF (ours)} & \textbf{xLSTMTime} & \textbf{PatchTST} & \textbf{minGRUTime} & \textbf{ELM} \\
\hline
\textbf{Metric} &  & \textbf{MSE\hspace{0.5cm}MAE} & \textbf{MSE\hspace{0.5cm}MAE} & \textbf{MSE\hspace{0.5cm}MAE} & \textbf{MSE\hspace{0.5cm}MAE} & \textbf{MSE\hspace{0.5cm}MAE} \\
\hline
\multirow{4}{*}{\textbf{PEMS03}} 
& 12 & \textbf{\textcolor{red}{0.061}}\hspace{0.5cm}\textbf{\textcolor{red}{0.162}} & \textbf{\textcolor{blue}{0.064}}\hspace{0.5cm}\textbf{\textcolor{blue}{0.166}} & 0.075\hspace{0.5cm}0.181 & 0.071\hspace{0.5cm}0.176 & 0.128\hspace{0.5cm}0.239 \\
& 24 & \textbf{\textcolor{red}{0.083}}\hspace{0.5cm}\textbf{\textcolor{blue}{0.190}} & \textbf{\textcolor{blue}{0.084}}\hspace{0.5cm}\textbf{\textcolor{red}{0.189}} & 0.107\hspace{0.5cm}0.216 & 0.103\hspace{0.5cm}0.212 & 0.315\hspace{0.5cm}0.414 \\
& 48 & \textbf{\textcolor{red}{0.127}}\hspace{0.5cm}\textbf{\textcolor{blue}{0.236}} & \textbf{\textcolor{blue}{0.132}}\hspace{0.5cm}\textbf{\textcolor{red}{0.235}} & 0.177\hspace{0.5cm}0.276 & 0.168\hspace{0.5cm}0.275 & 0.714\hspace{0.5cm}0.631 \\
& 96 & \textbf{\textcolor{red}{0.227}}\hspace{0.5cm}\textbf{\textcolor{red}{0.326}} & \textbf{\textcolor{blue}{0.257}}\hspace{0.5cm}\textbf{\textcolor{blue}{0.352}} & 0.346\hspace{0.5cm}0.398 & 0.280\hspace{0.5cm}0.361 & 1.703\hspace{0.5cm}1.035 \\
&Avg \% Improvement & \textbf{\textcolor{red}{7.46\%}}\hspace{0.5cm}\textbf{\textcolor{red}{2.97\%}} 
& \hspace{0.5cm}
& \hspace{0.5cm}
& \hspace{0.5cm}
& \hspace{0.5cm} \\

\hline
\multirow{4}{*}{\textbf{PEMS04}}  
& 12 & \textbf{\textcolor{red}{0.072}}\hspace{0.5cm}\textbf{\textcolor{red}{0.171}} & \textbf{\textcolor{red}{0.072}}\hspace{0.5cm}\textbf{\textcolor{blue}{0.173}} & \textbf{\textcolor{blue}{0.091}}\hspace{0.5cm}0.195 & 0.093\hspace{0.5cm}0.193 & 0.133\hspace{0.5cm}0.247 \\
& 24 & \textbf{\textcolor{red}{0.089}}\hspace{0.5cm}\textbf{\textcolor{red}{0.193}} & \textbf{\textcolor{blue}{0.093}}\hspace{0.5cm}\textbf{\textcolor{blue}{0.194}} & 0.140\hspace{0.5cm}0.245 & 0.124\hspace{0.5cm}0.232 & 0.247\hspace{0.5cm}0.338 \\
& 48 & \textbf{\textcolor{red}{0.114}}\hspace{0.5cm}\textbf{\textcolor{red}{0.222}} & \textbf{\textcolor{blue}{0.117}}\hspace{0.5cm}\textbf{\textcolor{blue}{0.225}} & 0.187\hspace{0.5cm}0.283 & 0.214\hspace{0.5cm}0.313 & 0.548\hspace{0.5cm}0.519 \\
& 96 & \textbf{\textcolor{red}{0.154}}\hspace{0.5cm}\textbf{\textcolor{red}{0.260}} & \textbf{\textcolor{blue}{0.160}}\hspace{0.5cm}\textbf{\textcolor{blue}{0.263}} & 0.278\hspace{0.5cm}0.352 & 0.284\hspace{0.5cm}0.370 & 1.062\hspace{0.5cm}0.766\\ & Avg \% Improvement & \textbf{\textcolor{red}{2.73\%}}\hspace{0.5cm}\textbf{\textcolor{red}{1.26\%}} 
& \hspace{0.5cm}
& \hspace{0.5cm} 
& \hspace{0.5cm}
& \hspace{0.5cm} \\

\hline
\multirow{4}{*}{\textbf{PEMS07}} 
& 12 & \textbf{\textcolor{red}{0.057}}\hspace{0.5cm}\textbf{\textcolor{red}{0.146}} & \textbf{\textcolor{blue}{0.058}}\hspace{0.5cm}\textbf{\textcolor{blue}{0.147}} & 0.070\hspace{0.5cm}0.166 & 0.278\hspace{0.5cm}0.352 & 0.160\hspace{0.5cm}0.263 \\
& 24 & \textbf{\textcolor{red}{0.072}}\hspace{0.5cm}\textbf{\textcolor{red}{0.166}} & \textbf{\textcolor{blue}{0.072}}\hspace{0.5cm}\textbf{\textcolor{blue}{0.167}} & 0.111\hspace{0.5cm}0.211 & 0.105\hspace{0.5cm}0.211 & 0.232\hspace{0.5cm}0.323 \\
& 48 & \textbf{\textcolor{red}{0.101}}\hspace{0.5cm}\textbf{\textcolor{red}{0.196}} & \textbf{\textcolor{blue}{0.101}}\hspace{0.5cm}\textbf{\textcolor{red}{0.196}} & 0.201\hspace{0.5cm}\textbf{\textcolor{blue}{ 0.291}} & 0.220\hspace{0.5cm}0.321 & 0.543\hspace{0.5cm}0.519 \\
& 96 & \textbf{\textcolor{red}{0.130}}\hspace{0.5cm}\textbf{\textcolor{red}{0.226}} & \textbf{\textcolor{red}{0.130}}\hspace{0.5cm}\textbf{\textcolor{blue}{0.227}} & 0.299\hspace{0.5cm}0.337 &\textbf{\textcolor{blue}{ 0.254}}\hspace{0.5cm}0.346 & 1.066\hspace{0.5cm}0.748 \\
&Avg \% Improvement& \textbf{\textcolor{red}{0.28\%}}\hspace{0.5cm}\textbf{\textcolor{red}{0.54\%}} 
& \hspace{0.5cm}
& \hspace{0.5cm} 
& \hspace{0.5cm} 
& \hspace{0.5cm} \\

\hline
\multirow{4}{*}{\textbf{PEMS08}}  
& 12 & \textbf{\textcolor{red}{0.070}}\hspace{0.5cm}\textbf{\textcolor{red}{0.166}} & \textbf{\textcolor{blue}{0.072}}\hspace{0.5cm}\textbf{\textcolor{blue}{0.169}} & 0.087\hspace{0.5cm}0.186 & 0.082\hspace{0.5cm}0.184 & 0.295\hspace{0.5cm}0.406 \\
& 24 & \textbf{\textcolor{red}{0.091}}\hspace{0.5cm}\textbf{\textcolor{red}{0.188}} & \textbf{\textcolor{blue}{0.097}}\hspace{0.5cm}\textbf{\textcolor{blue}{0.193}} & 0.124\hspace{0.5cm}0.222 & 0.121\hspace{0.5cm}0.222 & 0.233\hspace{0.5cm}0.322 \\
& 48 & \textbf{\textcolor{red}{0.140}}\hspace{0.5cm}\textbf{\textcolor{red}{0.228}} & \textbf{\textcolor{blue}{0.146}}\hspace{0.5cm}\textbf{\textcolor{blue}{0.232}} & 0.210\hspace{0.5cm}0.291 & 0.201\hspace{0.5cm}0.294 & 0.529\hspace{0.5cm}0.507 \\
& 96 & \textbf{\textcolor{red}{0.222}}\hspace{0.5cm}\textbf{\textcolor{red}{0.279}} & \textbf{\textcolor{blue}{0.231}}\hspace{0.5cm}\textbf{\textcolor{blue}{0.284}} & 0.376\hspace{0.5cm}0.393 & 0.351\hspace{0.5cm}0.392 & 1.109\hspace{0.5cm}0.754 \\
&Avg \% Improvement & \textbf{\textcolor{red}{4.78\%}}\hspace{0.5cm}\textbf{\textcolor{red}{1.83\%}} 
& \hspace{0.5cm} 
& \hspace{0.5cm}
& \hspace{0.5cm} 
& \hspace{0.5cm} \\

\hline
\end{tabular}%
}
\caption{Performance Comparison of EMTSF MoE Design with Component Models on the PEMS Dataset}
\label{tab:PEMSComparison1}
\end{table}

Some recent TSF models such as WPMixer \cite{murad2025wpmixer}, CycleNet \cite{lin2024cyclenet}, and Timeexer \cite{wang2024timexer} have proposed interesting TSF ideas. WPMixer is a novel MLP-based model that leverages the benefits of patching, multi-resolution wavelet decomposition, and mixing for TSF.  CycleNet uses the "Residual Cycle Forecasting" technique, which utilizes learnable recurrent cycles to model the inherent periodic patterns within sequences. TimeXer enhances the Transformer's capability to reconcile endogenous and exogenous information, where patch-wise self-attention and variate-wise cross-attention are used simultaneously. As can be seen from Table \ref{tab:ComparisonWithWpCycleTimeExer}, WPMixer is competitive with our EMTSF model for the ETT datasets, but lags in other datasets. This could be attributed to the Wavelet backbone being able to capture the hourly or five minute patterns of the transformer temperature in the ETTh/m datasets. WPMixer can be a good candidate model as one of the experts in our EMTSF framework.

\begin{table}[h]
\centering
\scalebox{0.72}{
\scriptsize
\setlength{\tabcolsep}{3pt}
\begin{tabular}{|c|c|c|c|c|c|c|}
\hline
\textbf{Models} & \textbf{Horizon} & \textbf{EMTSF ((Ours))} & \textbf{WPMixer } & \textbf{CycleNet /Linear} & \textbf{CycleNet /MLP} & \textbf{TimeXer} \\ \hline
\textbf{Metric} &                  & \textbf{MSE\hspace{0.5cm}MAE} & \textbf{MSE\hspace{0.5cm}MAE} & \textbf{MSE\hspace{0.5cm}MAE} & \textbf{MSE\hspace{0.5cm}MAE} & \textbf{MSE\hspace{0.5cm}MAE} \\ \hline

\textbf{Weather}   & 96  & \textcolor{red}{0.138}\hspace{0.5cm}\textcolor{red}{0.177} & 0.141\hspace{0.5cm}\textcolor{blue}{0.188} & 0.167\hspace{0.5cm}0.221 & \textcolor{blue}{0.140}\hspace{0.5cm}0.200 & 0.157\hspace{0.5cm}0.205 \\
                   & 192 & \textcolor{red}{0.182}\hspace{0.5cm}\textcolor{red}{0.220} & \textcolor{blue}{0.185}\hspace{0.5cm}\textcolor{blue}{0.229} & 0.212\hspace{0.5cm}0.258 & 0.190\hspace{0.5cm}0.240 & 0.204\hspace{0.5cm}0.247 \\
                   & 336 & \textcolor{red}{0.232}\hspace{0.5cm}\textcolor{red}{0.260} &\textcolor{blue}{ 0.236}\hspace{0.5cm}\textcolor{blue}{0.271} & 0.260\hspace{0.5cm}0.293 & 0.243\hspace{0.5cm}0.283 & 0.261\hspace{0.5cm}0.290 \\
                   & 720 & \textcolor{red}{0.305}\hspace{0.5cm}\textcolor{red}{0.315} &\textcolor{blue}{ 0.307\hspace{0.5cm}0.321} & 0.328\hspace{0.5cm}0.339 & 0.322\hspace{0.5cm}0.330 & 0.340\hspace{0.5cm}0.341 \\ \hline

\textbf{Traffic}   & 96  & \textcolor{red}{0.343\hspace{0.5cm}0.225} & \textcolor{blue}{0.354\hspace{0.5cm}0.246} & 0.397\hspace{0.5cm}0.278 & 0.386\hspace{0.5cm}0.268 & 0.428\hspace{0.5cm}0.271 \\
                   & 192 & \textcolor{red}{0.369}\hspace{0.5cm}\textcolor{red}{0.238} &\textcolor{blue}{ 0.371\hspace{0.5cm}0.253} & 0.411\hspace{0.5cm}0.283 & 0.404\hspace{0.5cm}0.276 & 0.448\hspace{0.5cm}0.282 \\
                   & 336 & \textcolor{red}{0.382}\hspace{0.5cm}\textcolor{red}{0.242} & \textcolor{blue}{0.387\hspace{0.5cm}0.267} & 0.424\hspace{0.5cm}0.289 & 0.416\hspace{0.5cm}0.281 & 0.473\hspace{0.5cm}0.289 \\
                   & 720 & \textcolor{blue}{0.424}\hspace{0.5cm}\textcolor{red}{0.270} & \textcolor{red}{0.403}\hspace{0.5cm}\textcolor{blue}{0.289} & 0.450\hspace{0.5cm}0.305 & 0.445\hspace{0.5cm}0.300 & 0.516\hspace{0.5cm}0.307 \\ \hline

\textbf{Electricity} & 96  & \textcolor{red}{0.126\hspace{0.5cm}0.217} & \textcolor{blue}{0.128}\hspace{0.5cm}0.222 &\textcolor{red}{0.126}\hspace{0.5cm}\textcolor{blue}{0.221} & \textcolor{red}{0.126}\hspace{0.5cm}0.221 & 0.140\hspace{0.5cm}0.242 \\
                     & 192 & \textcolor{red}{0.144}\hspace{0.5cm}\textcolor{red}{0.234} & 0.145\hspace{0.5cm}\textcolor{blue}{0.237} & \textcolor{red}{0.144}\hspace{0.5cm}\textcolor{blue}{0.237} & \textcolor{red}{0.144}\hspace{0.5cm}\textcolor{blue}{0.237} & 0.157\hspace{0.5cm}0.256 \\
                     & 336 & \textcolor{red}{0.158\hspace{0.5cm}0.248} & 0.161\hspace{0.5cm}0.256 & \textcolor{blue}{0.160\hspace{0.5cm}0.254} & \textcolor{blue}{0.160}\hspace{0.5cm}0.255 & 0.176\hspace{0.5cm}0.275 \\
                     & 720 & \textcolor{red}{0.190\hspace{0.5cm}0.277} &\textcolor{blue}{0.196}\hspace{0.5cm}\textcolor{blue}{0.287} & 0.198\hspace{0.5cm}\textcolor{blue}{0.287} & 0.199\hspace{0.5cm}0.291 & 0.211\hspace{0.5cm}0.306 \\ \hline

\textbf{ETTh1}     & 96  & \textcolor{blue}{0.359}\hspace{0.5cm}\textcolor{blue}{0.384} &\textcolor{red}{ 0.347}\hspace{0.5cm}\textcolor{red}{0.383} & 0.374\hspace{0.5cm}0.396 & 0.382\hspace{0.5cm}0.403 & 0.382\hspace{0.5cm}0.403 \\
                   & 192 & \textcolor{blue}{0.399\hspace{0.5cm}0.411} & \textcolor{red}{0.381\hspace{0.5cm}0.408} & 0.406\hspace{0.5cm}0.415 & 0.421\hspace{0.5cm}0.426 & 0.429\hspace{0.5cm}0.435 \\
                   & 336 & \textcolor{blue}{0.418\hspace{0.5cm}0.422} &\textcolor{red}{ 0.382\hspace{0.5cm}0.412 }& 0.431\hspace{0.5cm}0.430 & 0.449\hspace{0.5cm}0.444 & 0.468\hspace{0.5cm}0.448 \\
                   & 720 & \textcolor{blue}{0.436\hspace{0.5cm}0.454} & \textcolor{red}{0.405\hspace{0.5cm}0.432} & 0.450\hspace{0.5cm}0.464 & 0.497\hspace{0.5cm}0.485 & 0.469\hspace{0.5cm}0.461 \\ \hline

\textbf{ETTh2}     & 96  & \textcolor{blue}{0.262}\hspace{0.5cm}\textcolor{red}{0.324} & \textcolor{red}{0.253}\hspace{0.5cm}\textcolor{blue}{0.328} & 0.279\hspace{0.5cm}0.341 & 0.300\hspace{0.5cm}0.355 & 0.286\hspace{0.5cm}0.338 \\
                   & 192 & \textcolor{blue}{0.328\hspace{0.5cm}0.371} & \textcolor{red}{0.303\hspace{0.5cm}0.364} & 0.342\hspace{0.5cm}0.385 & 0.373\hspace{0.5cm}0.403 & 0.363\hspace{0.5cm}0.389 \\
                   & 336 & \textcolor{blue}{0.347\hspace{0.5cm}0.387} &\textcolor{red}{ 0.305\hspace{0.5cm}0.371} & 0.371\hspace{0.5cm}0.413 & 0.384\hspace{0.5cm}0.419 & 0.414\hspace{0.5cm}0.423 \\
                   & 720 & \textcolor{blue}{0.381}\hspace{0.5cm}\textcolor{red}{0.417} &\textcolor{red}{ 0.373\hspace{0.5cm}0.417} & 0.426\hspace{0.5cm}0.451 & 0.428\hspace{0.5cm}0.450 & 0.408\hspace{0.5cm}\textcolor{blue}{0.432 }\\ \hline

\textbf{ETTm1}     & 96  & \textcolor{red}{0.271\hspace{0.5cm}0.325} & \textcolor{blue}{0.275\hspace{0.5cm}0.333} & 0.299\hspace{0.5cm}0.348 & 0.297\hspace{0.5cm}0.351 & 0.318\hspace{0.5cm}0.356 \\
                   & 192 & \textcolor{blue}{0.322}\hspace{0.5cm}\textcolor{red}{0.351} &\textcolor{red}{ 0.319}\hspace{0.5cm}\textcolor{blue}{0.362} & 0.334\hspace{0.5cm}0.370 & 0.338\hspace{0.5cm}0.377 & 0.362\hspace{0.5cm}0.383 \\
                   & 336 & \textcolor{blue}{0.350}\hspace{0.5cm}\textcolor{red}{0.370 }& \textcolor{red}{0.347}\hspace{0.5cm}\textcolor{blue}{0.384} & 0.368\hspace{0.5cm}0.386 & 0.374\hspace{0.5cm}0.400 & 0.395\hspace{0.5cm}0.407 \\
                   & 720 & \textcolor{blue}{0.414}\hspace{0.5cm}\textcolor{red}{0.404} &\textcolor{red}{ 0.403}\hspace{0.5cm}\textcolor{blue}{0.414} & 0.417\hspace{0.5cm}0.414 & 0.436\hspace{0.5cm}0.431 & 0.452\hspace{0.5cm}0.441 \\ \hline

\textbf{ETTm2}     & 96  & \textcolor{red}{0.156\hspace{0.5cm}0.240} & \textcolor{blue}{0.159\hspace{0.5cm}0.246} & 0.159\hspace{0.5cm}0.247 & 0.178\hspace{0.5cm}0.262 & 0.171\hspace{0.5cm}0.256 \\
                   & 192 & \textcolor{red}{0.212\hspace{0.5cm}0.280} & \textcolor{blue}{0.214\hspace{0.5cm}0.286} & 0.226\hspace{0.5cm}0.287 & 0.238\hspace{0.5cm}0.303 & 0.237\hspace{0.5cm}0.299 \\
                   & 336 & \textcolor{red}{0.263\hspace{0.5cm}0.315} & \textcolor{blue}{0.266\hspace{0.5cm}0.322} & 0.292\hspace{0.5cm}0.322 & 0.292\hspace{0.5cm}0.339 & 0.296\hspace{0.5cm}0.338 \\
                   & 720 & \textcolor{blue}{0.351}\hspace{0.5cm}\textcolor{red}{0.371} & \textcolor{red}{0.344}\hspace{0.5cm}\textcolor{blue}{0.374} & 0.374\hspace{0.5cm}0.391 & 0.374\hspace{0.5cm}0.391 & 0.392\hspace{0.5cm}0.394 \\ \hline
\end{tabular}
}
\caption{Performance of EMTSF (ours), WPMixer, CycleNet, and TimeXer across various datasets and forecast horizons.}
\label{tab:ComparisonWithWpCycleTimeExer}
\end{table}


\section{Conclusion}
 Time series forecasting has been a challenging field due to its non-stationary nature, noise, seasonality, and unexpected events. In the TSF domain, there has been a difference in opinion as to whether more complex models result in better prediction, or simpler models should be used. For example, recent research has attempted use of LLM models for TSF with good success, e.g., Time-LLM \cite{jin2024time}, only to be refuted shortly by showing that removing the LLM layers in this model does not degrade performance. Recently, a large scale Transformer architecture (2.4 billion parameters) termed Time-MoE  \cite{shi2025time} reported impressive results surpassing current TSF models. 

In this work, we combine multiple small strong performing models in a mixture of experts transformer-based gating framework. We select the models that complement each other such that different aspects of TSF from seasonality, trend, short term as well long term pattern comprehension are cooperatively learnt and used in the prediction process. Our Transformer-based gating model controls the per time step contribution from each of the component models. In our MoE framework, we select an enhanced linear-based model, a Transformer-based model,  xLSTM, and minGRU based models. The xLSTM and minGRU are recent enhancements to the traditional LSTM and GRU and provide special benefits for long term forecasting in the TSF domain. Our MoE EMTSF model demonstrates better results on standard benchmarks as compared to both current state-of-the-art TSF models and MoE frameworks.

The EMTSF design developed in this work is easily extensible to adding more complementary experts in our MoE framework. For example, the recently proposed WPMixer\cite{murad2025wpmixer} based on the wavelet decomposition and mixing can be helpful in forecasting certain short duration cyclical patterns. Our future work is focused on drafting such experts, and in combining Soft MoE \cite{puigcerversparse} approach with our strong expert based MoE design.

\bibliography{ecai}

\end{document}